\documentclass[english]{article}
\UseRawInputEncoding
\usepackage{babel}
\usepackage{array}
\usepackage{float}
\usepackage{amsmath}
\usepackage{amsthm}
\usepackage{amssymb}
\usepackage{graphicx}
\PassOptionsToPackage{normalem}{ulem}
\usepackage{ulem}
\usepackage[]
 {hyperref}

\makeatletter

\providecommand{\tabularnewline}{\\}

\usepackage[preprint,nonatbib]{neurips_2026}
\usepackage[utf8]{inputenc} 
\usepackage[T1]{fontenc}    
\usepackage{hyperref}       
\usepackage{url}            
\usepackage{booktabs}       
\usepackage{amsfonts}       
\usepackage{nicefrac}       
\usepackage{microtype}      
\usepackage{xcolor}         

\usepackage{graphicx}
\usepackage[table]{xcolor}
\usepackage{longtable}

\definecolor{lightgray}{gray}{0.90} 

\title{Advantage Matching: A General Reinforcement Learning Framework for LLM Reasoning}

\author{%
  Kien Do \\
  VinSmart Future \\
  \texttt{v.kiendd5@vinsmartfuture.tech} \\
  \And
  Tu Nguyen \\
  FPT Smart Cloud \\
  \texttt{tuna37@fpt.com} \\
  \And
  Vu-An Hoang \\
  FPT Smart Cloud \\
  \texttt{anhv10@fpt.com} \\
  \And
  Cuong Pham \\
  PTIT \\
  \texttt{cuongpv@ptit.edu.vn} \\
  \And
  Trung Le \\
  Monash University \\
  \texttt{trunglm@monash.edu} \\
  \And
  Khoat Than \\
  SoICT, HUST \\
  \texttt{khoattq@soict.hust.edu.vn} \\
  \And
  Dieu Hoang Vu \\
  Phenikaa University \\
  \texttt{dieu.vuhoang@phenikaa-uni.edu.vn} \\
  \And
  Tuc Nguyen \\
  Indiana University \\
  \texttt{tucnguye@iu.edu} \\
}

\makeatother

\theoremstyle{plain}
\newtheorem{thm}{\protect\theoremname}
\newtheorem{prop}{\protect\propositionname}
\newtheorem{lem}{\protect\lemmaname}
\providecommand{\lemmaname}{Lemma}
\providecommand{\propositionname}{Proposition}
\providecommand{\theoremname}{Theorem}

\begin{document}
\maketitle

\global\long\def\Expect{\mathbb{E}}%
\global\long\def\Real{\mathbb{R}}%
\global\long\def\Data{\mathcal{D}}%
\global\long\def\Loss{\mathcal{L}}%
\global\long\def\Normal{\mathcal{N}}%
\global\long\def\softmax{\text{softmax}}%
\global\long\def\ELBO{\text{ELBO}}%
\global\long\def\argmin#1{\underset{#1}{\text{argmin}}}%
\global\long\def\argmax#1{\underset{#1}{\text{argmax}}}%
\global\long\def\refer{\text{ref}}%
\global\long\def\old{\text{old}}%

\begin{abstract}
Reinforcement learning (RL) has emerged as a promising paradigm for
improving reasoning in large language models (LLMs), enabling models
to optimize full reasoning trajectories using task-level feedback.
However, existing RL-based approaches suffer from several limitations,
including reliance on surrogate objectives and approximate advantage
estimators, token-level optimization that overlooks trajectory-level
dependencies, and predominantly on-policy training that restricts
the utilization of offline data. We introduce \textbf{Advantage Matching},
a general framework for KL-regularized RL that leverages the principled
relationship between the policy ratio and the soft advantage to learn
the optimal policy by aligning the two quantities. Our formulation
yields a family of stable and flexible objectives that support critic-free
optimization, trajectory-level reasoning, and seamless integration
of off-policy and offline data. Empirical results on diverse mathematical
reasoning and code generation benchmarks and model families show improvements
in performance of our method compared to existing baselines.
\footnote{This work was conducted when Kien was working at FPT Smart Cloud}
\end{abstract}

\section{Introduction}

Large language models (LLMs) have achieved remarkable success across
a broad range of tasks \cite{brown2020language,touvron2023llama,bai2023qwen,bi2024deepseek},
yet reliable multi-step reasoning remains a bottleneck. A natural
approach to address this challenge is to cast reasoning as a reinforcement
learning (RL) problem, where the model generates full reasoning trajectories
and improves them using task-level rewards as feedback. Unlike supervised
finetuning (SFT), which passively imitates fixed expert demonstrations,
this active learning paradigm allows models to iteratively refine
their own outputs, and has driven substantial progress in reasoning
performance \cite{guo2025deepseek,shao2025deepseekmath}.

Early RL-based approaches, such as RLHF \cite{ouyang2022training},
combine learned reward models with policy optimization algorithms
like PPO \cite{schulman2017proximal}. While effective, these methods
are often brittle in practice, requiring careful tuning and accurate
value estimation. More recently, critic-free methods such as Group
Relative Policy Optimization (GRPO) \cite{shao2024deepseekmath} have
gained traction by eliminating explicit value functions and instead
estimating advantages through relative comparisons among sampled responses.
These approaches significantly simplify training and have demonstrated
strong empirical results, particularly in reasoning-intensive domains.

However, despite their success, existing methods share several fundamental
limitations. First, they rely on surrogate objectives and approximate
advantage estimators, which can introduce bias and lead to suboptimal
policy updates \cite{ahmadian2024back,hu2025reinforce++,liu2025understanding}.
Second, most existing approaches perform optimization at the token
level, implicitly modeling reasoning as a sequence of local decisions
despite rewards being defined over complete trajectories \cite{zheng2025group}.
They typically propagate the same trajectory-level reward signal to
all tokens, overlooking the dependencies among reasoning steps and
limiting the model's flexibility to discover diverse, task-specific
reasoning strategies. Third, many methods are inherently on-policy,
requiring fresh samples from the current model at each iteration,
which constrains scalability and prevents effective use of large offline
datasets such as curated reasoning traces or expert demonstrations
\cite{liu2025dapo,yuan2025trajectory}.

These limitations highlight a fundamental mismatch between existing
policy optimization objectives and the trajectory-level nature of
reasoning. To address this issue, we propose \textbf{\emph{Advantage
Matching}}\textbf{ (AM)}, a general policy optimization framework
for KL-regularized RL that exploits the principled relationship between
the policy ratio and the soft advantage. We theoretically establish
that the optimal policy emerges as a fixed point when these two quantities
are matched, yielding a principled optimization objective without
relying on conventional surrogate objectives. For LLM reasoning, this
formulation naturally enables trajectory-level optimization, supports
critic-free estimation of soft advantages, and extends seamlessly
to off-policy and offline settings, allowing it to accomodates diverse
data sources while controlling the bias--variance trade-off. Extensive
experiments across diverse mathematical reasoning and code generation
benchmarks and model families demonstrate that AM consistently outperforms
existing RL-based approaches, validating its effectiveness and generalizability.

\section{Preliminary}

Consider the \emph{KL-regularized reinforcement learning (RL)} problem,
whose objective is to learn a policy $\pi$ that maximizes the expected
discounted KL-augmented return:
\begin{equation}
\mathcal{J}_{\beta}\left(\pi\right):=\Expect_{\tau\sim\pi}\left[\sum^{\infty}_{t=0}\gamma^{t}\left(r\left(s_{t},a_{t}\right)-\beta D_{\text{KL}}\left(\pi\left(\cdot|s_{t}\right)||\pi_{\text{ref}}\left(\cdot|s_{t}\right)\right)\right)\right]\label{eq:KL_reg_RL_objective}
\end{equation}
where $\pi_{\text{ref}}$ is a reference policy, $\tau=\left(s_{0},a_{0},s_{1},a_{1},...\right)$
denotes a trajectory generated by $\pi$, $\gamma\in(0,1]$ is the
discount factor, and $\beta>0$ controls the strength of the KL regularization.
The KL divergence term constrains the learned policy from deviating
excessively from the reference policy, thereby mitigating reward over-optimization
while preserving its desirable distributional characteristics.

For a policy $\pi$, we define the soft state value function $A_{\pi,\beta}$,
soft state-action value function $Q_{\pi,\beta}$, and soft advantage
function $A_{\pi,\beta}$ as:
\begin{align}
V_{\pi,\beta}\left(s\right) & :=\Expect_{\tau\sim\pi}\left[\sum^{\infty}_{t=0}\gamma^{t}\left(r\left(s_{t},a_{t}\right)-\beta\log\frac{\pi\left(a_{t}|s_{t}\right)}{\pi_{\text{ref}}\left(a_{t}|s_{t}\right)}\right)\Big|s_{0}=s\right]\label{eq:soft_value_func}\\
Q_{\pi,\beta}\left(s,a\right) & :=\Expect_{\tau\sim\pi}\left[r\left(s_{0},a_{0}\right)+\sum^{\infty}_{t=1}\gamma^{t}\left(r\left(s_{t},a_{t}\right)-\beta\log\frac{\pi\left(a_{t}|s_{t}\right)}{\pi_{\text{ref}}\left(a_{t}|s_{t}\right)}\right)\Big|s_{0}=s,a_{0}=a\right]\label{eq:soft_Q_value_func}\\
A_{\pi,\beta}\left(s,a\right) & :=Q_{\pi,\beta}\left(s,a\right)-V_{\pi,\beta}\left(s\right)\label{eq:soft_adv_func}
\end{align}
Since the per-state policy optimization problem induced by the KL-regularized
objective is strictly concave in $\pi\left(\cdot|s\right)$ for $\beta>0$,
it admits a \emph{unique optimal policy} $\pi^{*}$ with the following
closed-form expression (see Appendix~\ref{subsec:KL-Regularized-RL-Framework}):
\begin{align}
\pi^{*}\left(a|s\right) & =\frac{\pi_{\text{ref}}\left(a|s\right)\exp\left(\frac{1}{\beta}Q_{*,\beta}\left(s,a\right)\right)}{\int_{\mathcal{A}}\pi_{\text{ref}}\left(a'|s\right)\exp\left(\frac{1}{\beta}Q_{*,\beta}\left(s,a'\right)\right)da'}\label{eq:opt_policy_KL_reg_1}\\
 & =\pi_{\text{ref}}\left(a|s\right)\exp\left(\frac{1}{\beta}\left(Q_{*,\beta}\left(s,a\right)-V_{*,\beta}\left(s\right)\right)\right)\label{eq:opt_policy_KL_reg_2}
\end{align}
or equivalently,
\begin{equation}
\boxed{\pi^{*}\left(a|s\right)=\pi_{\text{ref}}\left(a|s\right)\exp\left(\frac{A_{*,\beta}\left(s,a\right)}{\beta}\right)}\label{eq:KL_opt_adv_identity}
\end{equation}
Here, $V_{*,\beta}\left(s\right):=\sup_{\pi}V_{\pi,\beta}\left(s\right)$
and $Q_{*,\beta}\left(s,a\right):=\sup_{\pi}Q_{\pi,\beta}\left(s,a\right)$
denote the optimal soft state value and soft state-action value functions,
respectively; and $A_{*,\beta}\left(s,a\right):=Q_{*,\beta}\left(s,a\right)-V_{*,\beta}\left(s\right)$
is the optimal soft advantage. $V_{*,\beta}$ and $Q_{*,\beta}$ are
linked through the soft Bellman optimality equation:
\begin{equation}
V_{*,\beta}\left(s\right)=\beta\log\int_{\mathcal{A}}\pi_{\text{ref}}\left(a|s\right)\exp\left(\frac{1}{\beta}Q_{*,\beta}\left(s,a\right)\right)da\label{eq:soft_Bellman_opt_eq_V_and_Q}
\end{equation}

\emph{Maximum Entropy (MaxEnt) RL} \cite{haarnoja2017reinforcement,nachum2017bridging,schulman2017equivalence,haarnoja2018soft}
is a special case of KL-regularized RL in which $\pi_{\text{ref}}$
is uniform. There, the identity reduces to (see Appendix.~\ref{subsec:Connection-to-MaxEnt}):
\begin{equation}
\boxed{\pi^{*}\left(a|s\right)=\exp\left(\frac{A_{*,\beta}\left(s,a\right)}{\beta}\right)}\label{eq:MaxEnt_opt_adv_identity}
\end{equation}

\section{Advantage Matching}

Although the identity in Eq.~\ref{eq:KL_opt_adv_identity} does not
directly yield the optimal policy $\pi^{*}$, as it depends on the
unknown $A_{*,\beta}$, it nevertheless provides a useful characterization
that can guide policy optimization. This insight is formalized in
the following theorem.
\begin{thm}
For the KL-regularized RL objective described in Eq.~\ref{eq:KL_reg_RL_objective},
the identity $\pi\left(a|s\right)=\pi_{\text{ref}}\left(a|s\right)\exp\left(\frac{A_{\pi,\beta}\left(s,a\right)}{\beta}\right)$
holds for all $\left(s,a\right)$ \textbf{if and only if} $\pi$ is
$\pi^{*}$.\label{Optimal-Policy-Characterization}
\end{thm}
\begin{proof}
See Appendix~\ref{subsec:Proof-of-Optimal-Policy-Characterization}.

Theorem 1 leverages the $\gamma$-contraction property of the soft
Bellman optimality operator (Eq.~{[}{]}) to establish that is $\pi^{*}$
the unique fixed point of the pointwise matching condition between
$\pi\left(a|s\right)$ and $\pi_{\text{ref}}\left(a|s\right)\exp\left(\frac{A_{\pi,\beta}\left(s,a\right)}{\beta}\right)$.
This characterization motivates our \textbf{\emph{Advantage Matching}}\textbf{
(AM)} objective, defined as follows:
\begin{equation}
\mathcal{L}_{\text{AM}}\left(\theta\right):=\Expect_{p\left(s,a\right)}\left[\frac{1}{2\beta}\left(\beta\log\frac{\pi_{\theta}\left(a|s\right)}{\pi_{\text{ref}}\left(a|s\right)}-A_{\pi_{\bar{\theta}},\beta}\left(s,a\right)\right)^{2}\right]\label{eq:AdvMatch}
\end{equation}

\[
\Expect_{p\left(s,a\right)}\left[\frac{1}{2}\left(\log\frac{\pi_{\theta}\left(a|s\right)}{\pi_{\text{ref}}\left(a|s\right)}-\frac{A_{\pi_{\bar{\theta}},\beta}\left(s,a\right)}{\beta}\right)^{2}\right]
\]

where $p\left(s,a\right)$ is a joint state-action distribution, and
$\pi_{\bar{\theta}}$ is $\pi_{\theta}$ \emph{with gradients stopped}.
We refer to the term $\beta\log\frac{\pi_{\theta}\left(a|s\right)}{\pi_{\text{ref}}\left(a|s\right)}-A_{\pi_{\bar{\theta}},\beta}\left(s,a\right)$
as the \emph{Advantage Matching (AM) error}.
\end{proof}

If $A_{\pi_{\bar{\theta}},\beta}\left(s,a\right)$ is estimated accurately,
minimizing $\mathcal{L}_{\text{AM}}\left(\theta\right)$ will encourage
$\pi_{\theta}$ to align with $\pi^{*}$. However, accurately estimating
$A_{\pi_{\bar{\theta}},\beta}\left(s,a\right)$ under a general $p\left(s,a\right)$
is challenging in practice. Moreover, the gradient update for a general
$p\left(s,a\right)$ does not guarantee improvement in $\mathcal{J}_{\beta}\left(\pi_{\theta}\right)$.

To enable reliable estimation of $A_{\pi_{\bar{\theta}},\beta}\left(s,a\right)$
and ensure monotonic improvement in $\mathcal{J}_{\beta}\left(\pi_{\theta}\right)$,
$p\left(s,a\right)$ must coincide with $d_{\pi_{\bar{\theta}},\gamma}\left(s\right)\pi_{\bar{\theta}}\left(a|s\right)$,
which leads to the \emph{on-policy} AM loss:
\begin{align}
\mathcal{L}^{\text{on-policy}}_{\text{AM}}\left(\theta\right) & =\Expect_{d_{\pi_{\bar{\theta}},\beta}\left(s\right)}\Expect_{\pi_{\bar{\theta}}\left(a|s\right)}\left[\frac{1}{2\beta\left(1-\gamma\right)}\left(\beta\log\frac{\pi_{\theta}\left(a|s\right)}{\pi_{\text{ref}}\left(a|s\right)}-A_{\pi_{\bar{\theta}},\beta}\left(s,a\right)\right)^{2}\right]\label{eq:AdvMatch_onpolicy_1}\\
 & =\Expect_{\tau\sim\pi_{\bar{\theta}}}\left[\sum^{\infty}_{t=0}\frac{\gamma^{t}}{2\beta}\left(\beta\log\frac{\pi_{\theta}\left(a_{t}|s_{t}\right)}{\pi_{\text{ref}}\left(a_{t}|s_{t}\right)}-A_{\pi_{\bar{\theta}},\beta}\left(s_{t},a_{t}\right)\right)^{2}\right]\label{eq:AdvMatch_onpolicy_2}
\end{align}

This choice of $p\left(s,a\right)$ is justified by the fact that
$\nabla_{\theta}\mathcal{L}^{\text{on-policy}}_{\text{AM}}\left(\theta\right)=-\nabla_{\theta}\mathcal{J}_{\beta}\left(\pi_{\theta}\right)$,
with a formal proof provided in Appendix~\ref{subsec:Gradient-Analysis}.

\subsection{Soft Advantage Approximation}

As in standard RL, the soft advantage $A_{\pi,\beta}\left(s,a\right)$
can be estimated using temporal-difference (TD) methods \cite{sutton1998reinforcement}.
In particular, the $n$-step look-ahead estimator is given by:
\begin{equation}
\hat{A}^{\left(n\right)}_{\pi,\beta}\left(s_{0},a_{0}\right)\approx r\left(s_{0},a_{0}\right)+\sum^{n-1}_{t=1}\gamma^{t}\left(r\left(s_{t},a_{t}\right)-\beta\log\frac{\pi\left(a_{t}|s_{t}\right)}{\pi_{\text{ref}}\left(a_{t}|s_{t}\right)}\right)+\gamma^{n}V_{\pi,\beta}\left(s_{n}\right)-V_{\pi,\beta}\left(s_{0}\right)
\end{equation}
with derivation provided in Appendix~\ref{subsec:Details-on-Estimating-the-Soft-Advantage}.

This estimator can be further generalized using TD($\lambda$) (with
$0<\lambda<1$) \cite{schulman2016high}, yielding:
\begin{equation}
\hat{A}^{\left(\lambda\right)}_{\pi,\beta}\left(s_{0},a_{0}\right):=\left(1-\lambda\right)\sum^{\infty}_{n=1}\lambda^{n-1}\hat{A}^{(n)}_{\pi,\beta}\left(s_{0},a_{0}\right)
\end{equation}

In practice, computing $\hat{A}^{\left(n\right)}_{\pi,\beta}$ requires
access to $V_{\pi,\beta}$, which is typically approximated by a parametric
critic $V_{\phi}$ trained to regress the KL-augmented return. Because
$V_{\phi}$ does not explicitly depend on $\pi$, it introduces bias
in approximating $V_{\pi,\beta}$, and consequently $\hat{A}^{\left(n\right)}_{\pi,\beta}$
and $\hat{A}^{\left(\lambda\right)}_{\pi,\beta}$ become biased estimators
of $A_{\pi,\beta}$, potentially degrading performance. Fortunately,
in the episodic setting, the soft advantage admits an \emph{unbiased},
\emph{critic-free} estimator by viewing the entire trajectory as a
single \emph{\textquotedblleft macro\textquotedblright{}} action.
We discuss this construction in detail in the next section.

\subsection{Trajectory-Level Advantage Matching}

Consider an episodic setting in which a trajectory $\tau=\left(a_{0},s_{1},a_{1},...,s_{T}\right)$
of horizon $T$ is generated by a policy $\pi$ from an initial state
$s_{0}$. We reinterpret this episode as a \emph{single-step} \emph{macro-transition}
by defining the initial macro state $\mathbf{s}_{0}=s_{0}$, the macro
action $\mathbf{a}_{0}=\left(a_{0},s_{1},a_{1},...,s_{T-1},a_{T-1}\right)$,
and the terminal macro state $\mathbf{s}_{1}=s_{T}$ reached from
$\mathbf{s}_{0}$ after executing $\mathbf{a}_{0}$.

To estimate $A_{\pi,\beta}\left(s_{0},\tau\right)\equiv A_{\pi,\beta}\left(\mathbf{s}_{0},\mathbf{a}_{0}\right)$,
we samples $M$ trajectories $\left\{ \tau^{\left(i\right)}\right\} ^{M}_{i=1}$
independently from $\pi\left(\cdot|s_{0}\right)$ and apply a leave-one-out
(LOO) Monte Carlo estimator \cite{kool2019buy,ahmadian2024back} as
follows:
\begin{equation}
A_{\pi,\beta}\left(s_{0},\tau^{(i)}\right)\approx\mathcal{R}\left(\tau^{(i)}\right)-\frac{1}{M-1}\sum_{j\neq i}\left(\mathcal{R}\left(\tau^{(j)}\right)-\beta\log\frac{\pi\left(\tau^{(j)}|s_{0}\right)}{\pi_{\text{ref}}\left(\tau^{(j)}|s_{0}\right)}\right)=\mathcal{R}\left(\tau^{(i)}\right)-B_{\pi_{\bar{\theta}}}\left(s_{0}\right)
\end{equation}
where $B_{\pi_{\bar{\theta}}}\left(s_{0}\right)$ is an action-independent
baseline defined by the LOO average.

Based on the above estimator of $A_{\pi,\beta}\left(s_{0},\tau^{(i)}\right)$,
we propose an \emph{unbiased} \emph{critic-free} \emph{trajectory-level
Advantage Matching (AM-Traj}) loss:
\begin{align}
\Loss^{\text{on-policy}}_{\text{AM-Traj}}\left(\theta\right) & =\Expect_{p\left(s_{0}\right)}\Expect_{\tau\sim\pi_{\bar{\theta}}\left(\cdot|s_{0}\right)}\left[\frac{1}{2\beta}\left(\beta\log\frac{\pi_{\theta}\left(\tau|s_{0}\right)}{\pi_{\text{ref}}\left(\tau|s_{0}\right)}-A_{\pi_{\bar{\theta}},\beta}\left(s_{0},\tau\right)\right)^{2}\right]\\
 & =\Expect_{p\left(\mathbf{s}_{0}\right)}\Expect_{\left\{ \tau^{(i)}\right\} ^{M}_{i=1}\sim\pi_{\bar{\theta}}\left(\cdot|s_{0}\right)}\left[\frac{1}{2\beta}\left(\beta\log\frac{\pi_{\theta}\left(\tau^{(i)}|s_{0}\right)}{\pi_{\text{ref}}\left(\tau^{(i)}|s_{0}\right)}-\mathcal{R}\left(\tau^{(i)}\right)+B_{\pi_{\bar{\theta}}}\left(s_{0}\right)\right)^{2}\right]
\end{align}

One can instead trade variance for simplicity by adopting baselines
that do not depend on $\pi_{\bar{\theta}}$ such as $B_{1}\left(s_{0}\right)=0$
or $B_{2}\left(s_{0}\right)=\frac{1}{M-1}\sum_{j\neq i}\mathcal{R}\left(\tau^{(j)}\right)$.
The former corresponds to directly matching $\beta\log\frac{\pi_{\theta}\left(\tau|s_{0}\right)}{\pi_{\text{ref}}\left(\tau|s_{0}\right)}$
to the soft Q-value.

\subsection{Extension to Off-policy and Offline RL\label{subsec:Extension-to-Off-policy-and-Offline-RL}}

In the \emph{off-policy} setting, trajectories $\tau$ are sampled
from a \emph{sampling }policy $\pi_{\text{s}}$ instead of the current
policy $\pi_{\bar{\theta}}$, yielding advantages $A_{\pi_{\text{s}},\beta}\left(\tau|s_{0}\right)$
rather than $A_{\pi_{\bar{\theta}},\beta}\left(\tau|s_{0}\right)$.
A straightforward way to mitigate sampling bias is to incorporate
importance weights $w_{\bar{\theta}}\left(s_{0},\tau\right)=\frac{\pi_{\bar{\theta}}\left(\tau|s_{0}\right)}{\pi_{\text{s}}\left(\tau|s_{0}\right)}$
into the objective:
\begin{equation}
\tilde{\Loss}^{\text{off-policy}}_{\text{AM-Traj}}\left(\theta\right)=\Expect_{p\left(s_{0}\right)}\Expect_{\tau\sim\pi_{\text{s}}\left(\cdot|s_{0}\right)}\left[\frac{1}{2\beta}w_{\bar{\theta}}\left(s_{0},\tau\right)\left(\beta\log\frac{\pi_{\theta}\left(\tau|s_{0}\right)}{\pi_{\text{ref}}\left(\tau|s_{0}\right)}-A_{\pi_{\text{s}},\beta}\left(s_{0},\tau\right)\right)^{2}\right]\label{eq:AM_Offpolicy_1}
\end{equation}
However, this formulation only compensates for the mismatch in the
sampling distribution $\tau\sim\pi_{\text{s}}\left(\cdot|s_{0}\right)$
while the advantage target $A_{\pi_{\text{s}},\beta}$ itself remains
biased. To better balance bias and variance, we introduce several
alternative off-policy objectives, described in Appendix~\ref{subsec:Different-Off-Policy-Variants}.
Among these variants, we find that the \emph{trajectory-level Hard
Advantage Matching (HAM-Traj)} objective performs particularly well
in practice:
\begin{equation}
\Loss^{\text{off-policy}}_{\text{AM-Traj}}\left(\theta\right)=\Expect_{p\left(s_{0}\right)}\Expect_{\tau\sim\pi_{\text{s}}\left(\cdot|s_{0}\right)}\left[\frac{1}{2\beta}w_{\bar{\theta}}\left(s_{0},\tau\right)\left(\beta\log\frac{\pi_{\theta}\left(\tau|s_{0}\right)}{\pi_{\text{ref}}\left(\tau|s_{0}\right)}-\left(\mathcal{R}\left(\tau\right)-\Expect_{\tau'\sim\pi_{\text{s}}\left(\cdot|s_{0}\right)}\left[\mathcal{R}\left(\tau'\right)\right]\right)\right)^{2}\right]\label{eq:AM_Offpolicy_2}
\end{equation}

Notably, the trajectory-level importance weight $w_{\bar{\theta}}\left(s_{0},\tau\right)=\frac{\pi_{\bar{\theta}}\left(\tau|s_{0}\right)}{\pi_{\text{s}}\left(\tau|s_{0}\right)}=\prod^{T-1}_{t=0}\frac{\pi_{\bar{\theta}}\left(a_{t}|s_{t}\right)}{\pi_{\text{s}}\left(a_{t}|s_{t}\right)}$
typically decreases with trajectory length, since $\frac{\pi_{\bar{\theta}}\left(a_{t}|s_{t}\right)}{\pi_{\text{s}}\left(a_{t}|s_{t}\right)}<1$
for most state-action pairs $\left(s_{t},a_{t}\right)$ in the trajectory
$\tau$ sampled from $\pi_{\text{s}}$. Consequently, the importance
weights naturally downweight excessively long trajectories. This behavior
implicitly regularizes the policy toward generating more concise reasoning
chains while still producing correct final answers. (see Appendix~\ref{subsec:Trajectory-Length-Dynamics})

Our AM framework also naturally extends to \emph{offline} RL. Let
$D^{*}=\left\{ \left(s_{0},\tau\right)\right\} $ denote a dataset
of expert demonstrations, where trajectories $\tau$ are assumed to
be sampled from the optimal policy $\pi^{*}$. We define the offline
trajectory-level AM loss (see Appendix~\ref{subsec:Performance-Difference-Analysis-for-KL-Regularized-RL}
for motivation) as:
\begin{align}
\Loss^{\text{offline}}_{\text{AM-Traj}}\left(\theta\right) & :=\Expect_{\left(s_{0},\cdot\right)\sim D^{*}}\Expect_{\tau\sim\pi_{\bar{\theta}}\left(\cdot|s_{0}\right)}\left[\left(\log\pi_{\theta}\left(\tau|s_{0}\right)-\log\pi^{*}\left(\tau|s_{0}\right)\right)^{2}\right]\\
 & =\Expect_{\left(s_{0},\tau\right)\sim D^{*}}\left[\frac{\pi_{\bar{\theta}}\left(\tau|s_{0}\right)}{\pi^{*}\left(\tau|s_{0}\right)}\left(\log\pi_{\theta}\left(\tau|s_{0}\right)-\log\pi^{*}\left(\tau|s_{0}\right)\right)^{2}\right]
\end{align}

Under the simplifying assumption that $\pi^{*}\left(\tau|s_{0}\right)\approx1$
for $\left(s_{0},\tau\right)\sim D^{*}$, this loss reduces to:
\begin{equation}
\Loss^{\text{offline}}_{\text{AM-Traj}}\left(\theta\right)\approx\Expect_{\left(s_{0},\tau\right)\sim D^{*}}\left[\pi_{\bar{\theta}}\left(\tau|s_{0}\right)\left(\log\pi_{\theta}\left(\tau|s_{0}\right)\right)^{2}\right]
\end{equation}

According to Propositions~\ref{KL_policy_minimization} and \ref{Optimal-Policy-Performance-Difference},
minimizing this loss is equivalent to minimizing the KL divergence
between $\pi_{\theta}$ and $\pi^{*}$ over trajectories, as well
as the performance gap $\mathcal{J}_{\beta}\left(\pi^{*}\right)-\mathcal{J}_{\beta}\left(\pi_{\theta}\right)$.
Since $\pi^{*}$ is globally optimal, reducing this gap guarantees
monotonic improvement in $\mathcal{J}_{\beta}\left(\pi_{\theta}\right)$.

Finally, we define the hybrid (online--offline) training objective
as:
\begin{equation}
\mathcal{L}^{\text{hybrid}}_{\text{AM-Traj}}\left(\theta\right):=\mathcal{L}^{\text{off-policy}}_{\text{AM-Traj}}\left(\theta\right)+\lambda_{\text{offline}}\ \mathcal{L}^{\text{offline}}_{\text{AM-Traj}}\left(\theta\right)
\end{equation}

\begin{prop}
Let $\pi_{\theta}$, $\pi'$ be two policies. Minimizing $\Expect_{\pi_{\bar{\theta}}\left(\cdot|s\right)}\left[\left(\log\pi_{\theta}\left(\cdot|s\right)-\log\pi'\left(\cdot|s\right)\right)^{2}\right]$
is equivalent to minimizing $D_{\text{KL}}\left(\pi_{\theta}\left(\cdot|s\right)\|\pi'\left(\cdot|s\right)\right)$
where $\pi_{\bar{\theta}}$ is $\pi_{\theta}$ with gradient stopped.\label{KL_policy_minimization}
\end{prop}
\begin{proof}
See Appendix~\ref{subsec:Proof-of-KL-Policy-Grad}.
\end{proof}

\begin{prop}
[Optimal Policy Performance Difference]The performance difference
between $\mathcal{J}_{\beta}\left(\pi^{*}\right)$ and $\mathcal{J}_{\beta}\left(\pi\right)$
has the following closed-form expression:\label{Optimal-Policy-Performance-Difference}
\[
\mathcal{D}_{\beta}\left(\pi^{*},\pi\right):=\mathcal{J}_{\beta}\left(\pi^{*}\right)-\mathcal{J}_{\beta}\left(\pi\right)=\frac{\beta}{1-\gamma}\Expect_{d_{\pi,\gamma}(s)}\left[D_{\text{KL}}\left(\pi\left(\cdot|s_{t}\right)\|\pi^{*}\left(\cdot|s_{t}\right)\right)\right]
\]
where $d_{\pi,\gamma}\left(s\right):=\left(1-\gamma\right)\sum^{\infty}_{t=0}\gamma^{t}p\left(s_{t}=s\right)$
is the discounted state visitation distribution.
\end{prop}
\begin{proof}
See Appendix~\ref{subsec:Performance-Difference-Analysis-for-KL-Regularized-RL}.
\end{proof}

\section{Related Work}

Reinforcement learning (RL) has become a standard paradigm for eliciting
reasoning in large language models (LLMs). Early approaches such as
RLHF \cite{ouyang2022training} train a reward model from human preference
data and optimize the policy via PPO \cite{schulman2017proximal}.
DPO \cite{rafailov2023direct} removes the need for an explicit reward
model by interpreting the log policy ratio $\beta\log\frac{\pi_{\theta}\left(\tau|s_{0}\right)}{\pi_{\text{ref}}\left(\tau|s_{0}\right)}$as
an implicit reward within a supervised preference-learning framework.
GRPO \cite{shao2024deepseekmath,guo2025deepseek} extends this critic-free
paradigm to more challenging settings such as long chain-of-thought
(CoT) reasoning by using Monte Carlo estimation of advantages, but
still operates at the token level and suffers from statistical biases.
Subsequent works address these issues through improved advantage estimation
(e.g., RLOO \cite{ahmadian2024back}, REINFORCE++ \cite{hu2025reinforce++}),
removal of length bias (Dr. GRPO \cite{liu2025understanding}), soft
trust-region updates (SAPO \cite{gao2025soft}), and various stabilization
techniques (DAPO \cite{yu2025dapo}), while other efforts extend optimization
to the trajectory level (e.g., GSPO \cite{zheng2025group}), which
is more natural given that rewards are typically defined at the completion
level \cite{li2024remax}.

Despite these advances, GRPO-based methods largely rely on online
sampling, limiting their ability to exploit offline expert data such
as ground-truth CoTs. To overcome this, several offline and hybrid
RL approaches have been proposed. DAPO \cite{liu2025dapo} introduces
an offline step-level objective related to advantage matching, replacing
$A_{\pi,\beta}$ with $A_{\pi_{\text{ref}},\beta}$. However, this
formulation does not guarantee monotonic improvement of the learned
policy $\pi_{\theta}$ in general settings. TBRM \cite{yuan2025trajectory}
treats the policy model logits as soft Q-values and minimizes the
squared loss derived from the soft Bellman optimality equation, while
OREO \cite{wang2025offline} jointly learns value and policy functions
using similar objectives inspired by path consistency learning (PCL)
\cite{nachum2017bridging}.

\section{Experiment}

\subsection{Experimental Setup}

\subsubsection{Datasets and Models}

\paragraph{Math reasoning}

We perform online RL-based finetuning using DeepMath-103k \cite{he2025deepmath}.
This dataset is primarily derived from Math StackExchange discussions
and undergoes an extensive curation and preprocessing pipeline to
convert the raw content into a high-quality question--answer format. 

For offline learning, we use the NuminaMath \cite{li2024numinamath}
dataset, which contains approximately 860k samples collected from
diverse sources spanning grade-school exercises to competition-level
mathematics. Each problem is paired with a concise and standardized
solution. We retain only samples whose solution length is below 4,000
tokens. Among the filtered samples, we preserve all AIME and AMC problems
due to their difficulty and rarity, as they constitute only about
0.4\% of the original dataset. Samples from the remaining sources
are randomly subsampled such that each source contributes approximately
3--15\% of the final dataset, resulting in a final corpus of roughly
128k samples.

We use the solutions provided in NuminaMath rather than those from
DeepMath-103k because the DeepMath-103k solutions are synthesized
using DeepSeek-R1, whereas many NuminaMath solutions are ground-truth
solutions extracted directly from raw PDF documents via OCR.

\paragraph{Code generation}

For code generation, we use TACO \cite{li2023taco}, a dataset of
algorithmic programming problems drawn from competitive programming
platforms. Its problems are more challenging than those in typical
code benchmarks and more closely resemble real-world programming scenarios.
Because the original dataset contains significantly more easy problems
than \textquotedblleft hard\textquotedblright{} or \textquotedblleft very
hard\textquotedblright{} ones, we first split it into two disjoint
subsets, an online dataset and an offline dataset, stratified by difficulty
level. We then filter each subset by difficulty level independently
so that all levels are represented in a balanced manner.

The online dataset is used for RL-based finetuning, which requires
only the problem statement and its test cases. The offline dataset
is restricted to problems with at least one provided solution. Since
these solutions are raw submissions from users of online coding platforms,
they follow no consistent format, structure, or input-output convention.
We therefore convert each solution into a unified function format
(\texttt{def solution()}) and verify its correctness by executing
it against the provided test inputs and expected outputs. For each
solution that passes verification, we generate a step-by-step explanation
using Gemini-3.5-Flash. Finally, we retain only samples whose problem
statement is shorter than 4,000 tokens and, for the offline dataset,
whose solution is also shorter than 4,000 tokens. This yields an online
dataset of 12k samples and an offline dataset of 4.8k samples, with
no problem shared between them.

\paragraph{Models}

We apply our method to finetune both general-purpose and domain-specialized
LLMs. For mathematical reasoning, we use \href{https://huggingface.co/meta-llama/Llama-3.1-8B-Instruct}{Llama 3.1 8B Instruct},
Qwen2.5-7B-Instruct and Qwen2.5-Math-7B \cite{Yang2024Qwen}. For
code generation, we use Qwen2.5-7B-Instruct and Qwen2.5-Coder-7B-Instruct
\cite{Hui2024Qwen}.

\subsubsection{Evaluation and Baselines}

We follow prior work \cite{shao2024deepseekmath,yu2025dapo,he2025deepmath}
and evaluate our method on six standard mathematical reasoning benchmarks:
\href{https://huggingface.co/datasets/zwhe99/aime90/viewer/default/2024}{AIME24},
\href{https://huggingface.co/datasets/math-ai/aime25}{AIME25}, \href{https://huggingface.co/datasets/zwhe99/amc23}{AMC23},
MATH-500 \cite{hendrycks2021measuring}, Minerva Math \cite{lewkowycz2022solving},
and OlympiadBench \cite{he2024olympiadbench}. For code generation,
we evaluate on five benchmarks: HumanEval \cite{chen2021evaluating},
HumanEval+ \cite{liu2023your}, MBPP \cite{austin2021program}, \href{https://huggingface.co/datasets/evalplus/mbppplus}{MBPP+}
\cite{liu2023your}, and LiveCodeBench \cite{jain2025livecodebench}.
HumanEval+ and MBPP+ extend their original counterparts with substantially
more test cases, enabling a more rigorous assessment of functional
correctness. LiveCodeBench collects problems from recent programming
contests, which reduces the risk of data contamination.

Performance is assessed using the Pass@$k$ metric, which measures
the probability that at least one of $k$ sampled solutions is correct.
An unbiased, numerically stable estimate of Pass@$k$ \cite{chen2021evaluating}
is given by: Pass@$k$ = $\Expect_{\text{problem}}\left[1-\frac{C\left(n-c,k\right)}{C\left(n,k\right)}\right]$
where $n\geq k$ denotes the number of generated samples and $c$
the number of correct ones. $n$ is set to 16 by default.

We compare our method with current critic-free online RL methods for
LLM reasoning including GRPO \cite{shao2024deepseekmath}, Dr.GRPO
\cite{liu2025understanding}, GSPO \cite{zheng2025group}, and TBRM
\cite{yuan2025trajectory}.

\subsubsection{Training Configurations}

\begin{table}
\begin{centering}
\resizebox{\textwidth}{!}{%
\begin{tabular}{lccccccc}
\hline 
Method & AIME24 & AIME25 & AMC23 & MATH-500 & Minerva & Olympiad & Avg.\tabularnewline
\hline 
\hline 
\textbf{Llama3.1-8B-Instruct} & 4.0 & 0.8 & 23.4 & 44.0 & 19.5 & 14.4 & 17.8\tabularnewline
+ GRPO & 5.6 & \textbf{2.9} & 26.7 & 47.8 & 25.0 & 15.8 & 20.6\tabularnewline
+ GSPO & 5.4 & 2.3 & 22.3 & 42.8 & 22.1 & 13.9 & 18.1\tabularnewline
+ Dr. GRPO & 6.0 & 1.7 & \uline{30.0} & 48.9 & \textbf{25.9} & 17.4 & 21.7\tabularnewline
\rowcolor{lightgray}+ AM (ours) & \uline{7.5} & \uline{2.5} & 29.1 & \uline{52.7} & \uline{25.8} & \uline{19.1} & \uline{22.8}\tabularnewline
\rowcolor{lightgray}+ AM-hybrid (ours) & \textbf{8.1} & 2.3 & \textbf{30.8} & \textbf{52.8} & 25.3 & \textbf{19.9} & \textbf{23.2}\tabularnewline
\hline 
\hline 
\textbf{Qwen2.5-7B-Instruct} & 11.5 & 6.9 & 53.0 & 75.8 & 34.0 & 33.2 & 35.7\tabularnewline
+ GRPO & \uline{14.4} & 11.5 & 53.6 & 77.0 & 36.6 & 39.8 & 38.8\tabularnewline
+ GSPO & 14.0 & 9.8 & 52.0 & 77.0 & \uline{37.3} & 39.5 & 38.3\tabularnewline
+ Dr. GRPO & 13.1 & 10.2 & 54.8 & 76.8 & \textbf{37.5} & 39.7 & 38.7\tabularnewline
\rowcolor{lightgray}+ AM (ours) & \textbf{14.8} & \textbf{13.3} & \uline{57.5} & \textbf{78.1} & \textbf{37.5} & \textbf{42.1} & \textbf{40.6}\tabularnewline
\rowcolor{lightgray}+ AM-hybrid (ours) & 13.8 & \uline{12.1} & \textbf{58.6} & \uline{77.8} & 36.2 & \uline{41.2} & \uline{40.0}\tabularnewline
\hline 
\hline 
\textbf{Qwen2.5-Math-7B} & 18.1 & 7.7 & 49.5 & 65.2 & 28.0 & 33.2 & 33.6\tabularnewline
+ GRPO & 18.3 & 10.8 & 58.1 & 80.1 & 34.7 & 39.6 & 40.2\tabularnewline
+ GSPO & 20.0 & 11.5 & 58.8 & 79.0 & 33.8 & 39.6 & 40.5\tabularnewline
+ Dr. GRPO & 19.8 & 11.9 & 60.3 & 80.3 & 34.6 & 40.4 & 41.2\tabularnewline
+ TBRM (offline) & 16.2 & 12.5 & 55.9 & 76.4 & 33.3 & 40.3 & 39.1\tabularnewline
+ TBRM (online) & \textbf{23.8} & 10.8 & \uline{64.5} & 77.3 & 34.1 & 39.6 & 41.0\tabularnewline
\rowcolor{lightgray}+ AM (ours) & \uline{22.3} & \uline{16.2} & 63.0 & \uline{81.1} & \uline{35.5} & \uline{41.1} & \uline{43.2}\tabularnewline
\rowcolor{lightgray}+ AM-hybrid (ours) & 19.8 & \textbf{16.7} & \textbf{67.3} & \textbf{84.1} & \textbf{38.2} & \textbf{45.5} & \textbf{45.3}\tabularnewline
\hline 
\end{tabular}}
\par\end{centering}
\caption{Pass@1 performance of different models finetuned using our method
and baseline online RL approaches across various mathematical benchmarks.
The best and second best results are highlighted in \textbf{bold}
and \uline{underlined}, respectively.\label{tab:main_results}}
\end{table}

We implement our method and baselines using the \href{https://huggingface.co/docs/trl/index}{Hugging Face TRL library}.
To accelerate training, a copy of the finetuned language model is
deployed with vLLM for generation. In this setup, the finetuned model
corresponds to the current policy $\pi_{\theta}$, while the vLLM-served
model acts as the sampling policy $\pi_{\text{s}}$, resulting in
an off-policy training setting. The parameters of the vLLM-served
model are periodically synchronized with those of the current model.
However, discrepancies in output probabilities still arise (see Appendix~\ref{subsec:Analysis-of-the-Importance-Weight}).
To address this, we employ the off-policy objective described in Eq.~\ref{eq:AM_Offpolicy_2}
for training. A detailed comparison with alternative off-policy objectives
is presented in Section.~\ref{subsec:Different-Off-Policy-Objectives}.

Following prior work \cite{shao2024deepseekmath}, we train with two
reward components: a format reward and an accuracy reward. The format
reward is set to 1 if the generated output adheres to the structure
\colorbox{lightgray}{<think>\textcolor{blue}{{[}Solution{]}}</think><answer>\textbackslash boxed\{\textcolor{blue}{{[}Final
answer{]}}\}</answer>}, and 0 otherwise. The format and accuracy
rewards are combined with weights of 1/3 and 2/3, respectively.

For mathematical reasoning experiments, the accuracy reward is assigned
a value of 1 when a valid answer can be extracted from the output
and matches the ground-truth answer. The answer-matching function
is adapted from the \href{https://github.com/zwhe99/DeepMath}{official implementation}
of the DeepMath-103k paper. For code generation experiments, the system
prompt instructs the model to place its final solution in a self-contained
\texttt{def solution()} function within a \texttt{python} code
block. The accuracy reward is set to 1 if the code block can be successfully
extracted and the resulting solution passes all test cases, and 0
otherwise. To evaluate correctness, the extracted solution is embedded
in a test script and executed against the test inputs and expected
outputs provided with each problem.

\begin{figure}
\centering{}\includegraphics[width=1\textwidth]{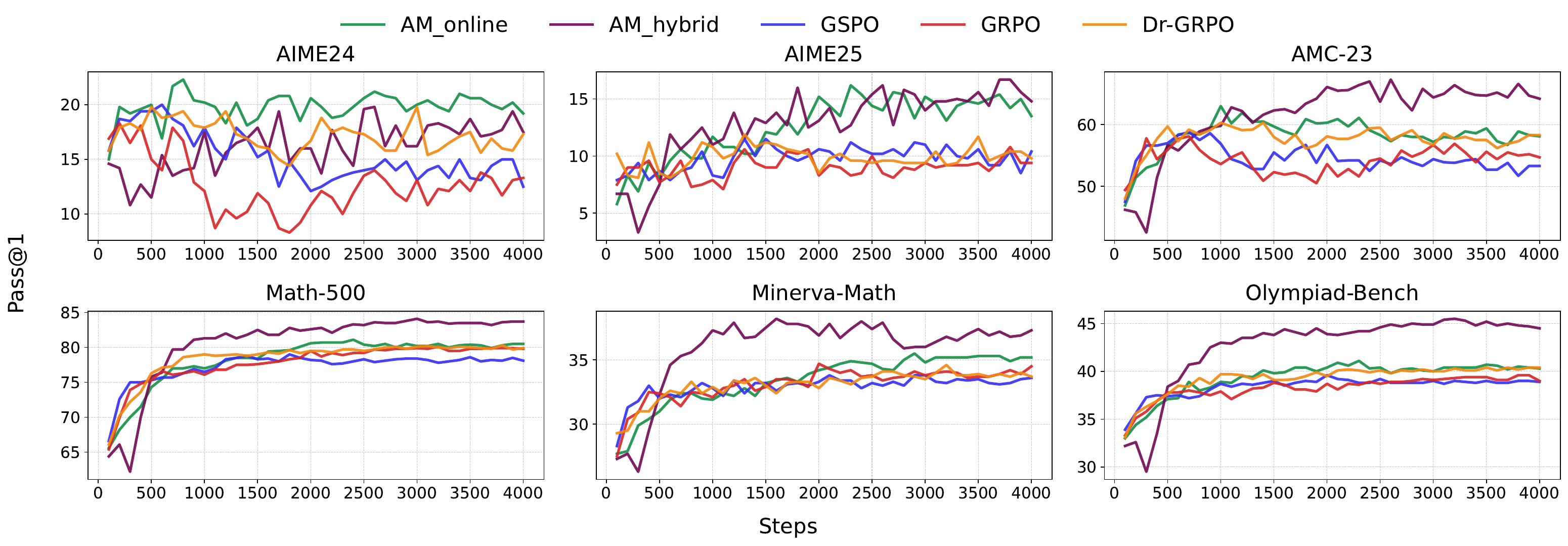}\caption{Pass@1 performance curves of Qwen2.5-Math-7B finetuned with our method
and other RL baselines across multiple benchmarks over the course
of training.\label{fig:Pass-1-curves-Qwen-Math}}
\end{figure}

Unless otherwise specified, we employ LoRA \cite{Hu2022lora} with
rank 64, scaling factor $\alpha$ = 128, and dropout rate 0.05. Models
are finetuned on either DeepMath-103k (math reasoning) or TACO (code
generation) for 4,000 iterations using the AdamW optimizer. We use
a learning rate of 2e-5 and a warmup ratio of 0.05 for Qwen models,
and a constant learning rate of 1e-5 for Llama3.1-8B-Instruct. The
total batch size is 32, with 16 generated responses per query, and
a maximum response length of 3072 tokens. The KL-regularization coefficient
$\beta$ is set to 0.001. These hyperparameters are shared across
all methods except TBRM, for which we follow the optimal configuration
specified in the original paper and use its official implementation.
Empirically, we find that TBRM can be finetuned stably only with Qwen2.5-Math-7B
for mathematical reasoning, whereas finetuning with other base models
leads to training collapse despite exploring various configurations.
This limitation is also reflected in the original TBRM paper, which
reports results exclusively for Qwen2.5-Math-7B. Accordingly, we report
TBRM results only for this base model.

For hybrid (online-offline) learning, we implement a new data loader
that, at each step, draws one sample from the online dataset and one
random sample from the offline dataset to form an online-offline pair.
For mathematical reasoning, the online and offline samples come from
DeepMath-103k and the filtered NuminaMath dataset, respectively. For
code generation, they come from the online and offline splits of TACO.
$\beta_{\text{offline}}$ is set to 2e-5 and decrease according to
the power rule ${0.992}^{\text{step}}$. In contrast, the coefficient
of the online accuracy reward increases linearly from 0 at the beginning
of training to $2/3$ at iteration 1k, after which it remains fixed. 

Our training procedure requires approximately 50 - 70 NVIDIA H100
GPU hours, while evaluation across all checkpoints and benchmarks
requires an additional 30 - 40 GPU hours, depending on the experimental
configuration.

\subsection{Main Results}

\begin{table}
\begin{centering}
\resizebox{\textwidth}{!}{%
\begin{tabular}{lcccccc}
\hline 
Method & HumanEval & HumanEval+ & MBPP & MBPP+ & LiveCodeBench & Avg.\tabularnewline
\hline 
\hline 
\textbf{Qwen2.5-7B-Instruct} & 75.3 & 69.2 & 69.4 & 65.5 & 24.2 & 60.7\tabularnewline
+ GRPO & 80.6 & 74.5 & 74,.9 & \uline{71.1} & 39.6 & 68.1\tabularnewline
+ GSPO & 80.1 & 75.1 & 73.8 & 70.0 & 40.2 & 67.8\tabularnewline
+ Dr. GRPO & 81.1 & 75.4 & \textbf{75.8} & \textbf{71.2} & 39.7 & 68.6\tabularnewline
\rowcolor{lightgray}+ AM (ours) & \uline{83.7} & \uline{77.6} & 74.9 & \uline{71.1} & \textbf{41.4} & \uline{69.7}\tabularnewline
\rowcolor{lightgray}+ AM-hybrid (ours) & \textbf{84.9} & \textbf{78.8} & \uline{75.1} & \textbf{71.2} & \uline{40.9} & \textbf{70.2}\tabularnewline
\hline 
\hline 
\textbf{Qwen2.5-Coder-7B-Instruct} & 81.6 & 75.4 & 70.2 & 65.2 & 26.3 & 63.7\tabularnewline
+ GRPO & \uline{88.0} & \textbf{83.4} & \uline{77.4} & \uline{72.9} & 42.5 & 72.8\tabularnewline
+ GSPO & 86.8 & 82.2 & 76.8 & 71.6 & 43.3 & 72.1\tabularnewline
+ Dr. GRPO & \uline{88.0} & 81.6 & 76.9 & 72.2 & 43.3 & 72.4\tabularnewline
\rowcolor{lightgray}+ AM (ours) & 87.9 & \uline{83.0} & \textbf{78.1} & \textbf{73.8} & \uline{45.6} & \textbf{73.7}\tabularnewline
\rowcolor{lightgray}+ AM-hybrid (ours) & \textbf{88.3} & 82.4 & 77.3 & 72.3 & \textbf{46.2} & \uline{73.3}\tabularnewline
\hline 
\end{tabular}}
\par\end{centering}
\caption{Pass@1 performance of different models finetuned using our method
and baseline online RL approaches across various code generation benchmarks.
The best and second best results are highlighted in \textbf{bold}
and \uline{underlined}, respectively.\label{tab:main_results-code}}
\end{table}

\paragraph{Math reasoning}

As shown in Table~\ref{tab:main_results}, AM consistently outperforms
competing baselines across most mathematical reasoning benchmarks.
Specifically, with Qwen2.5-7B-Instruct as the backbone, AM improves
Pass@1 by 1.8\% over GRPO. In the case the backbone is Qwen2.5-Math-7B,
our method achieves approximately 2\% higher Pass@1 than both Dr.
GRPO and TBRM. These improvements are also consistent across Pass@$k$
with $k\in\left\{ 2,4,8\right\} $ (see Appendix~\ref{subsec:PassAtK-Results}).

Incorporating additional offline data further improves the performance
of Advantage Matching (AM) when training Qwen2.5-Math-7B, yielding
approximately a 2\% increase in Pass@1 relative to the purely online
variant. In contrast, for Llama3.1-8B-Instruct and Qwen2.5-7B-Instruct,
the performance gap between AM and AM-hybrid is marginal, with AM-hybrid
even underperforming AM on several benchmarks. We conjecture that
the reasoning trajectories produced by math-specialized models are
more closely aligned with the distribution of solutions in the offline
dataset than those generated by general-purpose instruction models.
This distributional alignment likely reduces the optimization conflict
between online and offline objectives during joint training.

To further investigate the training dynamics, we present the performance
curves of Qwen2.5-7B-Instruct in Figure~\ref{fig:Pass-1-curves-Qwen-Math},
while the corresponding curves for Qwen2.5-Math-7B and Llama 3.1 8B
Instruct are deferred to Appendix~\ref{subsec:Additional-Performance-Curves}.
Despite some stochastic fluctuations across individual checkpoints,
our method consistently demonstrates a stronger upward trend compared
to the baselines. For AM-hybrid, we observe an initial decline in
performance during the early stages of training, followed by rapid
recovery and subsequent acceleration. We attribute this behavior to
the model initially prioritizing alignment with offline solutions
before effectively learning to optimize rewards on online samples.
Once the model adapts to the offline data distribution and successfully
extracts useful reasoning patterns -- particularly for math-specialized
models -- its performance improves substantially and eventually surpasses
all baseline methods.

\paragraph{Code generation}

Beyond mathematical reasoning, our method also demonstrates effectiveness
in code generation, achieving Pass@1 improvements of approximately
1\%--4\% across different experimental settings. For instance, with
Qwen2.5-Coder-7B-Instruct as the backbone, AM and AM-hybrid outperform
Dr. GRPO by 2.3\% and 2.9\%, respectively, on the challenging LiveCodeBench
benchmark.

\subsection{Ablation Studies}

\subsubsection{Different off-policy objectives\label{subsec:Different-Off-Policy-Objectives}}

We train different variants of our method using the off-policy objectives
described in Appendix~\ref{subsec:Different-Off-Policy-Variants}
and report the results in Table~\ref{tab:off_policy_results_abl}.
Among these variants, HAM-Traj achieves the best overall performance,
attaining an average Pass@1 score of 43.2\%. Owing to its favorable
bias--variance trade-off, HAM-Traj delivers strong gains on challenging
reasoning benchmarks such as AIME25 and AMC23, where it outperforms
other variants by a substantial margin. AM4SP-Traj, which introduces
higher bias than HAM-Traj, achieves the second-best overall performance.
In contrast, although RM-Traj performs well on MATH-500, Minerva Math,
and Olympiad Bench, it exhibits noticeably weaker results on more
difficult reasoning benchmarks including AIME24, AIME25, and AMC23.
These results suggest that the high variance induced by RM-Traj negatively
affects RL optimization in complex mathematical reasoning tasks.

\begin{table}
\begin{centering}
\resizebox{\textwidth}{!}{%
\begin{tabular}{lccccccc}
\hline 
Off-policy Objective & AIME24 & AIME25 & AMC23 & MATH-500 & Minerva & Olympiad & Avg.\tabularnewline
\hline 
\hline 
Naive (Eq.~\ref{eq:AM_Offpolicy_1}) & 22.1 & 11.9 & 60.3 & 78.9 & 34.8 & 41.1 & 41.5\tabularnewline
RM-Traj (Eq.~\ref{eq:RM_Traj_loss}) & 16.7 & 11.0 & 59.8 & 81.0 & \textbf{37.5} & \textbf{42.6} & 41.4\tabularnewline
AM4SP-Traj (Eq.~\ref{eq:AM4SP_Traj_loss}) & \textbf{22.5} & 15.0 & 61.7 & 80.0 & 34.2 & 41.2 & 42.4\tabularnewline
\rowcolor{lightgray}HAM-Traj (Eq.~\ref{eq:HAM_Traj_loss}) & 22.3 & \textbf{16.2} & \textbf{63.0} & \textbf{81.1} & 35.5 & 41.1 & \textbf{43.2}\tabularnewline
\hline 
\end{tabular}}
\par\end{centering}
\caption{Pass@1 performance of our method under different off-policy learning
objectives. The base model is Qwen2.5-Math-7B. The best results are
highlighted in \textbf{bold}. The gray-shaded row is reproduced from
Table~\ref{tab:main_results}. \label{tab:off_policy_results_abl}}
\end{table}

\subsubsection{Coefficient for the offline loss}

\begin{figure}
\begin{centering}
\includegraphics[width=0.85\textwidth]{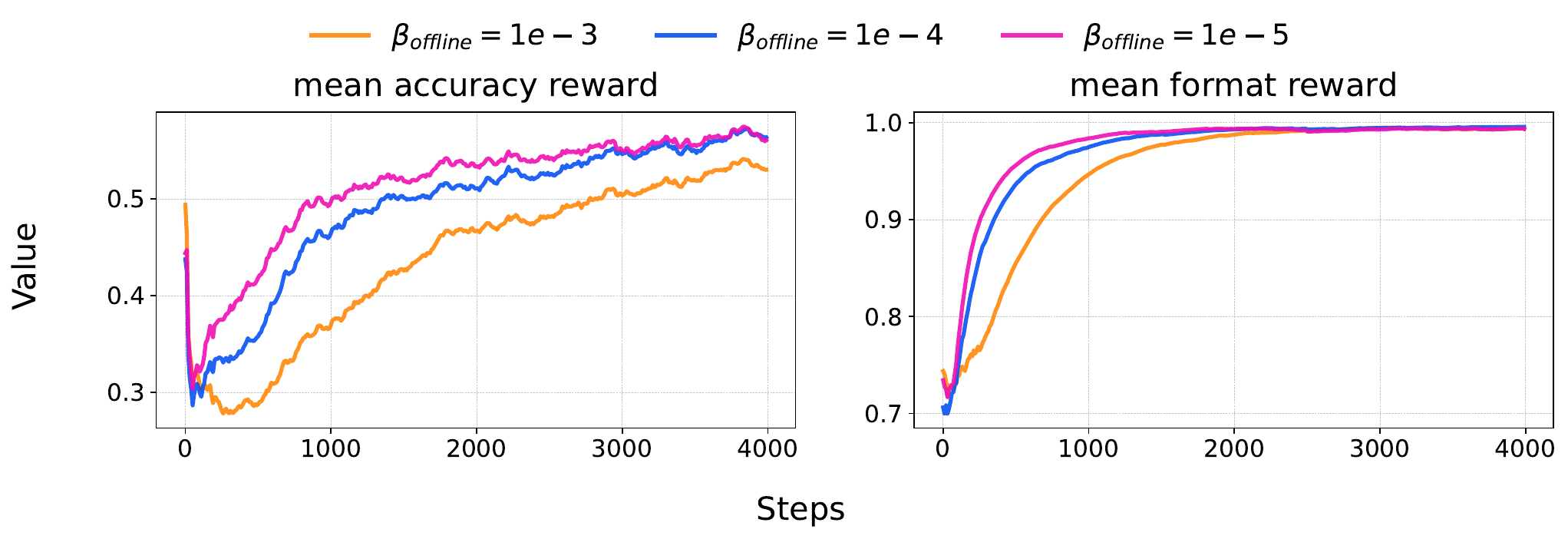}
\par\end{centering}
\caption{Smoothed accuracy and format reward curves of AM-hybrid under different
$\beta_{\text{offline}}$ values. The base model is Qwen2.5-7B-Instruct.\label{fig:beta_offline_abl}}
\end{figure}

We train multiple variants of AM-hybrid using different values of
$\beta_{\text{offline}}$ in $\left\{ {10}^{-3},{10}^{-4},{10}^{-5}\right\} $
and visualize the smoothed accuracy and format reward curves in Figure~\ref{fig:beta_offline_abl}.
We observe that when $\beta_{\text{offline}}$ is relatively large
(e.g., $1e-3$), the model places more emphasis on imitating offline
solutions rather than optimizing the online accuracy and format rewards,
leading to suboptimal performance. As $\beta_{\text{offline}}$ decreases,
the conflict between the trajectory distributions induced by offline
and online learning is reduced, enabling the model to learn a more
effective policy.

\section{Conclusion}

We introduced Advantage Matching (AM), a general KL-regularized RL
framework for LLM reasoning that learns the optimal policy through
a principled relationship between policy ratios and soft advantages.
We further derived trajectory-level, off-policy, and offline extensions
of AM, and demonstrated both the effectiveness and soundness of our
approach through consistent improvements over existing RL-based reasoning
baselines on challenging reasoning benchmarks.

Our work nevertheless has several limitations. First, our experiments
are restricted to mathematical reasoning with LLMs, and it remains
an important direction to extend AM to broader settings such as VLM,
VLA, and general control problems. Second, although AM is closely
connected to broader variational optimization frameworks, including
stochastic optimal control and score matching in diffusion models,
we leave a detailed theoretical analysis of these connections to future
work. Finally, our investigation of hybrid online--offline learning
remains preliminary, relying on a simple combination of online and
offline objectives and comparisons only against the online variant
of our method. Developing more principled strategies for integrating
online and offline RL under distribution mismatch remains an important
direction for future research.


\newpage{}

\appendix

\section{Mathematical Derivation}

\subsection{KL-Regularized RL Framework\label{subsec:KL-Regularized-RL-Framework}}

Recall that the KL-regularized RL problem involves maximizing the
following objective:
\begin{equation}
\mathcal{J}_{\beta}\left(\pi\right):=\Expect_{\tau\sim\pi}\left[\sum^{\infty}_{t=0}\gamma^{t}\left(r\left(s_{t},a_{t}\right)-\beta D_{\text{KL}}\left(\pi\left(\cdot|s_{t}\right)||\pi_{\text{ref}}\left(\cdot|s_{t}\right)\right)\right)\right]\label{eq:KL_reg_RL_obj_Appdx}
\end{equation}
where $\tau:=\left(s_{0},a_{0},s_{1},a_{1},...\right)$ denotes a
trajectory sampled from $\pi$.

Since $D_{\text{KL}}\left(\pi\left(\cdot|s_{t}\right)||\pi_{\text{ref}}\left(\cdot|s_{t}\right)\right)=\Expect_{\pi\left(a_{t}|s_{t}\right)}\left[\log\frac{\pi\left(a_{t}|s_{t}\right)}{\pi_{\text{ref}}\left(a_{t}|s_{t}\right)}\right]$,
using linearity of expectation gives:
\begin{equation}
\mathcal{J}_{\beta}\left(\pi\right)=\Expect_{\tau\sim\pi}\left[\sum^{\infty}_{t=0}\gamma^{t}\left(r\left(s_{t},a_{t}\right)-\beta\log\frac{\pi\left(a_{t}|s_{t}\right)}{\pi_{\text{ref}}\left(a_{t}|s_{t}\right)}\right)\right]\label{eq:KL_reg_RL_obj_Appdx_2}
\end{equation}

We define the \emph{soft value function} and \emph{soft state-action
value function} associated with the KL-regularized objective as:
\begin{align}
V_{\pi,\beta}\left(s\right) & :=\Expect_{\tau\sim\pi}\left[\sum^{\infty}_{t=0}\gamma^{t}\left(r\left(s_{t},a_{t}\right)-\beta\log\frac{\pi\left(a_{t}|s_{t}\right)}{\pi_{\text{ref}}\left(a_{t}|s_{t}\right)}\right)\Big|s_{0}=s\right]\\
Q_{\pi,\beta}\left(s,a\right) & :=r\left(s,a\right)+\Expect_{\tau\sim\pi}\left[\sum^{\infty}_{t=1}\gamma^{t}\left(r\left(s_{t},a_{t}\right)-\beta\log\frac{\pi\left(a_{t}|s_{t}\right)}{\pi_{\text{ref}}\left(a_{t}|s_{t}\right)}\right)\Big|s_{0}=s,a_{0}=a\right]
\end{align}

Assume that the environment reward $r\left(s,a\right)$ is deterministic
for each state-action pair. The corresponding Bellman expectation
equations are:
\begin{align}
V_{\pi,\beta}\left(s\right) & =\Expect_{\pi\left(a|s\right)}\left[Q_{\pi,\beta}\left(s,a\right)-\beta\log\frac{\pi\left(a|s\right)}{\pi_{\text{ref}}\left(a|s\right)}\right]\label{eq:soft_Bellman_eq_V}\\
Q_{\pi,\beta}\left(s,a\right) & =r\left(s,a\right)+\gamma\Expect_{p\left(s'|s,a\right)}\left[V_{\pi,\beta}\left(s'\right)\right]\label{eq:soft_Bellman_eq_Q}
\end{align}

Equivalently, combining the two equations yields:
\begin{align}
V_{\pi,\beta}\left(s\right) & =\Expect_{\pi\left(a|s\right)}\left[r\left(s,a\right)-\beta\log\frac{\pi\left(a|s\right)}{\pi_{\text{ref}}\left(a|s\right)}+\gamma\Expect_{p\left(s'|s,a\right)}\left[V_{\pi,\beta}\left(s'\right)\right]\right]\\
Q_{\pi,\beta}\left(s,a\right) & =r\left(s,a\right)+\gamma\Expect_{p\left(s'|s,a\right)}\Expect_{\pi\left(a'|s'\right)}\left[Q_{\pi,\beta}\left(s',a'\right)-\beta\log\frac{\pi\left(a'|s'\right)}{\pi_{\text{ref}}\left(a'|s'\right)}\right]
\end{align}

The \emph{soft advantage function} is defined as:
\begin{equation}
A_{\pi,\beta}\left(s,a\right):=Q_{\pi,\beta}\left(s,a\right)-V_{\pi,\beta}\left(s\right)
\end{equation}

Using the Bellman expectation equation for $V_{\pi,\beta}\left(s\right)$,
we obtain:
\begin{align}
A_{\pi,\beta}\left(s,a\right) & =Q_{\pi,\beta}\left(s,a\right)-\Expect_{\pi\left(a'|s\right)}\left[Q_{\pi,\beta}\left(s,a'\right)-\beta\log\frac{\pi\left(a'|s\right)}{\pi_{\text{ref}}\left(a'|s\right)}\right]\\
 & =Q_{\pi,\beta}\left(s,a\right)-\Expect_{\pi\left(a'|s\right)}\left[Q_{\pi,\beta}\left(s,a'\right)\right]+\beta D_{\text{KL}}\left(\pi\left(\cdot|s\right)\|\pi_{\text{ref}}\left(\cdot|s\right)\right)
\end{align}

The optimal soft value function is defined as:
\begin{align}
V_{*,\beta}\left(s\right) & :=\sup_{\pi}V_{\pi,\beta}\left(s\right)
\end{align}

Using the Bellman form of $V_{\pi,\beta}\left(s\right)$, this can
be written as a per-state optimization problem:
\begin{align}
V_{*,\beta}\left(s\right) & =\sup_{\pi}\Expect_{\pi\left(a|s\right)}\left[Q_{\pi,\beta}\left(s,a\right)-\beta\log\frac{\pi\left(a|s\right)}{\pi_{\text{ref}}\left(a|s\right)}\right]\\
 & =\sup_{\pi}\left(\left\langle \pi\left(\cdot|s\right),Q_{\pi,\beta}\left(s,\cdot\right)\right\rangle -\beta D_{\text{KL}}\left(\pi\left(\cdot|s\right)\|\pi_{\text{ref}}\left(\cdot|s\right)\right)\right)
\end{align}
where the supremum is taken over all probability distributions $\pi\left(\cdot|s\right)$
over action.

We therefore consider the per-state optimization problem:
\begin{equation}
\max_{\pi\left(\cdot|s\right)}\left(\left\langle \pi\left(\cdot|s\right),Q_{\pi,\beta}\left(s,\cdot\right)\right\rangle -\beta D_{\text{KL}}\left(\pi\left(\cdot|s\right)\|\pi_{\text{ref}}\left(\cdot|s\right)\right)\right)\ \ \text{s.t.}\ \ \int\pi\left(a|s\right)da=1
\end{equation}

This objective is strictly concave in $\pi\left(\cdot|s\right)$ due
to the negative KL term, and the feasible set is convex. Hence, a
unique optimal policy exists. Moreover, at the optimum, the functional
derivative of the Lagrangian w.r.t. $\pi\left(a|s\right)$ must vanish
for all $a$.

The Lagrangian is given by:
\begin{equation}
L:=\int\pi\left(a|s\right)\left(Q_{\pi,\beta}\left(s,a\right)-\beta\log\frac{\pi\left(a|s\right)}{\pi_{\text{ref}}\left(a|s\right)}\right)da+\lambda\left(\int\pi\left(a|s\right)da-1\right)
\end{equation}

Its functional derivative w.r.t. $\pi\left(a|s\right)$ is:
\begin{equation}
\frac{\partial L}{\partial\pi\left(a|s\right)}=Q_{\pi,\beta}\left(s,a\right)-\beta\left(\log\frac{\pi\left(a|s\right)}{\pi_{\text{ref}}\left(a|s\right)}+1\right)+\lambda\left(s\right)
\end{equation}

Setting $\frac{\partial L}{\partial\pi\left(a|s\right)}=0$ for $\pi^{*}\left(a|s\right)$,
we have:
\begin{equation}
\log\frac{\pi^{*}\left(a|s\right)}{\pi_{\text{ref}}\left(a|s\right)}=\frac{1}{\beta}\left(Q_{*,\beta}\left(s,a\right)+\lambda\left(s\right)-\beta\right)
\end{equation}

Exponentiating both sides yields:
\begin{equation}
\pi^{*}\left(a|s\right)=\pi_{\text{ref}}\left(a|s\right)\exp\left(\frac{1}{\beta}Q_{*,\beta}\left(s,a\right)\right)\exp\left(\frac{\lambda\left(s\right)-\beta}{\beta}\right)\label{eq:opt_policy_lagrange}
\end{equation}

The last exponential term is independent of $a$ and serves as a normalization
constant.

Imposing the constraint $\int\pi^{*}\left(a|s\right)da=1$, we obtain:
\begin{align}
\exp\left(\frac{\lambda\left(s\right)-\beta}{\beta}\right) & =\left(\int\pi_{\text{ref}}\left(a|s\right)\exp\left(\frac{1}{\beta}Q_{*,\beta}\left(s,a\right)\right)da\right)^{-1}
\end{align}

Substituting this expression back into Eq.~\ref{eq:opt_policy_lagrange}
gives the Boltzmann distribution form of the optimal policy:
\begin{equation}
\boxed{\pi^{*}\left(a|s\right)=\frac{\pi_{\text{ref}}\left(a|s\right)\exp\left(\frac{1}{\beta}Q_{*,\beta}\left(s,a\right)\right)}{\int\pi_{\text{ref}}\left(a|s\right)\exp\left(\frac{1}{\beta}Q_{*,\beta}\left(s,a\right)\right)da}}\label{eq:opt_policy_via_soft_opt_Q}
\end{equation}

Applying the soft Bellman expectation equation (Eq.~\ref{eq:soft_Bellman_eq_V})
to the optimal policy $\pi^{*}$ and using the above expression, we
obtain the soft Bellman optimality equation:
\begin{equation}
\boxed{V_{*,\beta}\left(s\right)=\beta\log\int\pi_{\text{ref}}\left(a|s\right)\exp\left(\frac{1}{\beta}Q_{*,\beta}\left(s,a\right)\right)da}\label{eq:soft_Bellman_optimality_equation_V}
\end{equation}

Equivalently,
\begin{equation}
\exp\left(\frac{1}{\beta}V_{*,\beta}\left(s\right)\right)=\int\pi_{\text{ref}}\left(a|s\right)\exp\left(\frac{1}{\beta}Q_{*,\beta}\left(s,a\right)\right)da\label{eq:normalization_constant}
\end{equation}

Substituting this result back into the normalization term of Eq.~\ref{eq:opt_policy_via_soft_opt_Q},
we have:
\begin{align}
\pi^{*}\left(a|s\right) & =\frac{\pi_{\text{ref}}\left(a|s\right)\exp\left(\frac{1}{\beta}Q_{*,\beta}\left(s,a\right)\right)}{\exp\left(\frac{1}{\beta}V_{*,\beta}\left(s\right)\right)}\\
 & =\pi_{\text{ref}}\left(a|s\right)\exp\left(\frac{1}{\beta}\left(Q_{*,\beta}\left(s,a\right)-V_{*,\beta}\left(s\right)\right)\right)\\
 & =\pi_{\text{ref}}\left(a|s\right)\exp\left(\frac{A_{*,\beta}\left(s,a\right)}{\beta}\right)
\end{align}

\subsection{Connection to MaxEnt RL\label{subsec:Connection-to-MaxEnt}}

In the special case where $\pi_{\text{ref}}$ is uniform over the
action space $\mathcal{A}$, the KL divergence satisfies:
\begin{equation}
D_{\text{KL}}\left(\pi\left(\cdot|s\right)||\pi_{\text{ref}}\left(\cdot|s\right)\right)=-\mathcal{H}\left(\pi\left(\cdot|s\right)\right)+\log\left|\mathcal{A}\right|
\end{equation}
where $\mathcal{H}\left(\pi\left(\cdot|s_{t}\right)\right)$ denotes
the Shannon entropy of $\pi\left(\cdot|s\right)$. Since $\log\left|\mathcal{A}\right|$
does not depend on $\pi$, the KL-regularized RL objective in Eq.~\ref{eq:KL_reg_RL_objective}
reduces (up to an additive constant) to the maximum entropy RL objective
\cite{nachum2017bridging,haarnoja2017reinforcement,haarnoja2018soft}:
\begin{equation}
\mathcal{J}_{\beta}\left(\pi\right):=\Expect_{\tau\sim\pi}\left[\sum^{\infty}_{t=0}\gamma^{t}\left(r\left(s_{t},a_{t}\right)+\beta\mathcal{H}\left(\pi\left(\cdot|s_{t}\right)\right)\right)\right]
\end{equation}

For this objective, $V_{*,\beta}\left(s\right)$ and $Q_{*,\beta}\left(s,a\right)$
satisfy the soft Bellman optimality equation:
\begin{equation}
V_{*,\beta}\left(s\right)=\beta\log\int_{\mathcal{A}}\exp\left(\frac{1}{\beta}Q_{*,\beta}\left(s,a\right)\right)da
\end{equation}
and the corresponding optimal policy is given by:
\begin{equation}
\pi^{*}\left(a|s\right)=\frac{\exp\left(\frac{1}{\beta}Q_{*,\beta}\left(s,a\right)\right)}{\int_{\mathcal{A}}\exp\left(\frac{1}{\beta}Q_{*,\beta}\left(s,a'\right)\right)da'}=\exp\left(\frac{1}{\beta}A_{*,\beta}\left(s,a\right)\right)
\end{equation}

\subsection{Proof of Theorem \ref{Optimal-Policy-Characterization}\label{subsec:Proof-of-Optimal-Policy-Characterization}}

From Eq.~\ref{eq:KL_opt_adv_identity}, it follows directly that
if $\pi$ is $\pi^{*}$ then $\pi\left(a|s\right)=\pi_{\text{ref}}\left(a|s\right)\exp\left(\frac{A_{\pi,\beta}\left(s,a\right)}{\beta}\right)$
for all $(s,a)$. Therefore, it suffices to prove that if
\[
\pi\left(a|s\right)=\pi_{\text{ref}}\left(a|s\right)\exp\left(\frac{A_{\pi,\beta}\left(s,a\right)}{\beta}\right)\ \ \forall\left(s,a\right)
\]
 then $\pi$ is optimal, i.e. $\pi$ is $\pi^{*}$. This condition
can be equivalently rewritten as:
\begin{align*}
 & \pi\left(a|s\right)=\pi_{\text{ref}}\left(a|s\right)\exp\left(\frac{A_{\pi,\beta}\left(s,a\right)}{\beta}\right)\\
\Leftrightarrow\  & \pi\left(a|s\right)=\pi_{\text{ref}}\left(a|s\right)\exp\left(\frac{1}{\beta}\left(Q_{\pi,\beta}\left(s,a\right)-V_{\pi,\beta}\left(s\right)\right)\right)\\
\Leftrightarrow\  & \pi\left(a|s\right)=\frac{\pi_{\text{ref}}\left(a|s\right)\exp\left(\frac{1}{\beta}Q_{\pi,\beta}\left(s,a\right)\right)}{\exp\left(\frac{1}{\beta}V_{\pi,\beta}\left(s\right)\right)}
\end{align*}

Since $\exp\left(\frac{1}{\beta}V_{\pi,\beta}\left(s\right)\right)$
does not depend on $a$, it serves as the normalization constant ensuring
that $\int_{a}\pi\left(a|s\right)da=1$. Consequently, we obtain:
\[
\exp\left(\frac{1}{\beta}V_{\pi,\beta}\left(s\right)\right)=\int\pi_{\text{ref}}\left(a|s\right)\exp\left(\frac{1}{\beta}Q_{\pi,\beta}\left(s,a\right)\right)da
\]
which is equivalent to:
\[
V_{\pi,\beta}\left(s\right)=\beta\log\int\pi_{\text{ref}}\left(a|s\right)\exp\left(\frac{1}{\beta}Q_{\pi,\beta}\left(s,a\right)\right)da
\]

Using the soft Bellman expectation equation for $Q_{\pi,\beta}\left(s,a\right)$
(Eq.~\ref{eq:soft_Bellman_eq_Q}), we have:
\[
V_{\pi,\beta}\left(s\right)=\beta\log\int\pi_{\text{ref}}\left(a|s\right)\exp\left(\frac{1}{\beta}\left(r\left(s,a\right)+\gamma\Expect_{p\left(s'|s,a\right)}\left[V_{\pi,\beta}\left(s'\right)\right]\right)\right)da
\]
which can be written compactly as:
\begin{equation}
\left(\mathcal{T}_{*,\beta}V_{\pi,\beta}\right)\left(s\right)=V_{\pi,\beta}\left(s\right)
\end{equation}

Here,$\mathcal{T}_{*,\beta}$ is the \emph{soft Bellman optimality
operator}, defined as:
\begin{equation}
\left(\mathcal{T}_{*,\beta}V\right)\left(s\right):=\beta\log\int\pi_{\text{ref}}\left(a|s\right)\exp\left(\frac{1}{\beta}\left(r\left(s,a\right)+\gamma\Expect_{p\left(s'|s,a\right)}\left[V\left(s'\right)\right]\right)\right)da
\end{equation}
In the following, we show that $\mathcal{T}_{*,\beta}$ is $\gamma$-contractive
under the supremum norm $\left\Vert \cdot\right\Vert _{\infty}$,
which implies the uniqueness of its fixed point and hence establishes
the optimality of $\pi$. This result relies on the following lemmas.
\begin{lem}
Let $f,g:\mathcal{A}\rightarrow\Real$, and define $F\left(f\right):=\beta\log\int\pi_{\text{ref}}\left(a|s\right)\exp\left(\frac{1}{\beta}f\left(a\right)\right)da$.
Then, \label{Bellman_optimality_op}
\[
\left|F\left(f\right)-F\left(g\right)\right|\leq\sup_{a}\left|f\left(a\right)-g\left(a\right)\right|
\]
\end{lem}
\begin{proof}
Let $\Delta:=\sup_{a}\left|f\left(a\right)-g\left(a\right)\right|$.
Then for all $a$,
\begin{align*}
 & f\left(a\right)-g\left(a\right)\leq\Delta\\
\Rightarrow & f\left(a\right)\leq g\left(a\right)+\Delta\\
\Rightarrow & \exp\left(\frac{1}{\beta}f\left(a\right)\right)\leq\exp\left(\frac{1}{\beta}f\left(a\right)\right)\exp\left(\frac{1}{\beta}\Delta\right)\\
\Rightarrow & \pi_{\text{ref}}\left(a|s\right)\exp\left(\frac{1}{\beta}f\left(a\right)\right)\leq\pi_{\text{ref}}\left(a|s\right)\exp\left(\frac{1}{\beta}f\left(a\right)\right)\exp\left(\frac{1}{\beta}\Delta\right)
\end{align*}

Integrating both sides over $a$ and then taking logarithms, we obtain:
\begin{align*}
 & \int\pi_{\text{ref}}\left(a|s\right)\exp\left(\frac{1}{\beta}f\left(a\right)\right)da\leq\int\pi_{\text{ref}}\left(a|s\right)\exp\left(\frac{1}{\beta}f\left(a\right)\right)\exp\left(\frac{1}{\beta}\Delta\right)da\\
\Rightarrow & \int\pi_{\text{ref}}\left(a|s\right)\exp\left(\frac{1}{\beta}f\left(a\right)\right)da\leq\exp\left(\frac{1}{\beta}\Delta\right)\int\pi_{\text{ref}}\left(a|s\right)\exp\left(\frac{1}{\beta}f\left(a\right)\right)da\\
\Rightarrow & \beta\log\int\pi_{\text{ref}}\left(a|s\right)\exp\left(\frac{1}{\beta}f\left(a\right)\right)da\leq\beta\log\left(\exp\left(\frac{1}{\beta}\Delta\right)\int\pi_{\text{ref}}\left(a|s\right)\exp\left(\frac{1}{\beta}f\left(a\right)\right)da\right)\\
\Rightarrow & \beta\log\int\pi_{\text{ref}}\left(a|s\right)\exp\left(\frac{1}{\beta}f\left(a\right)\right)da\leq\Delta+\beta\int\pi_{\text{ref}}\left(a|s\right)\exp\left(\frac{1}{\beta}f\left(a\right)\right)da\\
\Rightarrow & F\left(f\right)\leq F\left(g\right)+\Delta\\
\Rightarrow & F\left(f\right)-F\left(g\right)\leq\Delta
\end{align*}

By symmetry (swapping $f$ and $g$), we also have $F\left(g\right)-F\left(f\right)\leq\Delta$.
Therefore,
\begin{equation}
\boxed{\left|F\left(f\right)-F\left(g\right)\right|\leq\sup_{a}\left|f\left(a\right)-g\left(a\right)\right|}
\end{equation}
\end{proof}

\begin{lem}
Let $V_{1}\left(s\right)$, $V_{2}\left(s\right)$ be two value functions,
and define the corresponding state-action value function:\label{value_func_ineq}
\[
Q_{i}\left(s,a\right)=r\left(s,a\right)+\gamma\Expect_{p\left(s'|s,a\right)}\left[V_{i}\left(s'\right)\right],\ \ i\in\left\{ 1,2\right\} 
\]
Then,
\[
\sup_{a}\left|Q_{1}\left(s,a\right)-Q_{2}\left(s,a\right)\right|\leq\gamma\left\Vert V_{1}-V_{2}\right\Vert _{\infty}
\]
\end{lem}
\begin{proof}
For \emph{any} state-action pair $\left(s,a\right)$,
\begin{align*}
\left|Q_{1}\left(s,a\right)-Q_{2}\left(s,a\right)\right| & =\gamma\left|\Expect_{p\left(s'|s,a\right)}\left[V_{1}\left(s'\right)-V_{2}\left(s'\right)\right]\right|\\
 & \leq\gamma\Expect_{p\left(s'|s,a\right)}\left|V_{1}\left(s'\right)-V_{2}\left(s'\right)\right|\\
 & \leq\gamma\left\Vert V_{1}-V_{2}\right\Vert _{\infty}
\end{align*}

Taking supremum over action $a$, we have:
\begin{equation}
\boxed{\sup_{a}\left|Q_{1}\left(s,a\right)-Q_{2}\left(s,a\right)\right|\leq\gamma\left\Vert V_{1}-V_{2}\right\Vert _{\infty}}
\end{equation}
\end{proof}

Using Lemma~\ref{Bellman_optimality_op} and Lemma~\ref{value_func_ineq}
gives:
\begin{align}
\left|\left(\mathcal{T}_{*,\beta}V_{1}\right)\left(s\right)-\left(\mathcal{T}_{*,\beta}V_{2}\right)\left(s\right)\right| & \leq\sup_{a}\left|Q_{1}\left(s,a\right)-Q_{2}\left(s,a\right)\right|\\
 & \leq\gamma\left\Vert V_{1}-V_{2}\right\Vert _{\infty}
\end{align}
which holds for \emph{any} state $s$. Taking supremum over $s$,
we obtain:
\begin{equation}
\boxed{\left\Vert \mathcal{T}_{*,\beta}V_{1}-\mathcal{T}_{*,\beta}V_{2}\right\Vert _{\infty}\leq\gamma\left\Vert V_{1}-V_{2}\right\Vert _{\infty}}
\end{equation}

Thus, $\mathcal{T}_{*,\beta}$ is $\gamma$-contractive under the
supremum norm. 

According to the Banach fixed-point theorem, the equation $\mathcal{T}_{*,\beta}$
admits a \emph{unique} fixed point. Since both $V_{\pi,\beta}$ and
$V_{\pi^{*},\beta}$ satisfy $\left(\mathcal{T}_{*,\beta}V\right)\left(s\right)=V\left(s\right)$,
it follows that $V_{\pi,\beta}=V_{\pi^{*},\beta}$, for all states
$s$. This implies that $\pi$ is $\pi^{*}$.

\subsection{Details on Estimating the Soft Advantage\label{subsec:Details-on-Estimating-the-Soft-Advantage}}

To estimate $A_{\pi,\beta}\left(s,a\right)$, we bootstrap the soft
Q-value $Q_{\pi,\beta}\left(s,a\right)$ using temporal-difference
(TD) methods. Based on its definition, the $n$-step look ahead estimator
of $Q_{\pi,\beta}\left(s_{0},a_{0}\right)$ is given by:
\[
\hat{Q}^{(n)}_{\pi,\beta}\left(s_{0},a_{0}\right)=r\left(s_{0},a_{0}\right)+\sum^{n-1}_{t=1}\gamma^{t}\left(r\left(s_{t},a_{t}\right)-\beta\log\frac{\pi\left(a_{t}|s_{t}\right)}{\pi_{\text{ref}}\left(a_{t}|s_{t}\right)}\right)+\gamma^{n}V_{\pi,\beta}\left(s_{n}\right)
\]

The corresponding estimator for the soft advantage is then obtained
by subtracting the value baseline:
\begin{align*}
\hat{A}^{(n)}_{\pi,\beta}\left(s_{0},a_{0}\right) & :=\hat{Q}^{(n)}_{\pi,\beta}\left(s_{0},a_{0}\right)-V_{\pi,\beta}\left(s_{0}\right)
\end{align*}
which yields:
\[
\boxed{\hat{A}^{(n)}_{\pi,\beta}\left(s_{0},a_{0}\right)=r\left(s_{0},a_{0}\right)+\sum^{n-1}_{t=1}\gamma^{t}\left(r\left(s_{t},a_{t}\right)-\beta\log\frac{\pi\left(a_{t}|s_{t}\right)}{\pi_{\text{ref}}\left(a_{t}|s_{t}\right)}\right)+\gamma^{n}V_{\pi,\beta}\left(s_{n}\right)-V_{\pi,\beta}\left(s_{0}\right)}
\]

\subsection{Performance Difference Analysis for KL-Regularized RL\label{subsec:Performance-Difference-Analysis-for-KL-Regularized-RL}}

For any policy $\pi\in\Pi$ and value function $V:\mathcal{S}\rightarrow\mathbb{R}$,
we define the \emph{soft Bellman operator} $\mathcal{T}_{\pi,\beta}$:
\begin{equation}
\left(\mathcal{T}_{\pi,\beta}V\right)\left(s\right):=\Expect_{\pi\left(a|s\right)}\left[r\left(s,a\right)-\beta\log\frac{\pi\left(a|s\right)}{\pi_{\text{ref}}\left(a|s\right)}+\gamma\Expect_{p\left(s'|s,a\right)}\left[V\left(s'\right)\right]\right]\label{eq:soft_Bellman_operator}
\end{equation}

Let $V_{\pi,\beta}\left(s_{0}\right)$ and $V_{\pi',\beta}\left(s_{0}\right)$
denote the corresponding soft value functions of policies $\pi$ and
$\pi'$, respectively. For any initial state $s_{0}$, their difference
can be decomposed as:
\begin{align*}
 & V_{\pi,\beta}\left(s_{0}\right)-V_{\pi',\beta}\left(s_{0}\right)\\
=\  & V_{\pi,\beta}\left(s_{0}\right)-\left(\mathcal{T}_{\pi,\beta}V_{\pi',\beta}\right)\left(s_{0}\right)+\left(\mathcal{T}_{\pi,\beta}V_{\pi',\beta}\right)\left(s_{0}\right)-V_{\pi',\beta}\left(s_{0}\right)\\
=\  & \left(\left(\mathcal{T}_{\pi,\beta}V_{\pi,\beta}\right)\left(s_{0}\right)-\left(\mathcal{T}_{\pi,\beta}V_{\pi',\beta}\right)\left(s_{0}\right)\right)+\left(\left(\mathcal{T}_{\pi,\beta}V_{\pi',\beta}\right)\left(s_{0}\right)-V_{\pi',\beta}\left(s_{0}\right)\right)\\
=\  & \gamma\Expect_{\pi\left(a_{0}|s_{0}\right)}\Expect_{p\left(s_{1}|s_{0},a_{0}\right)}\left[V_{\pi,\beta}\left(s_{1}\right)-V_{\pi',\beta}\left(s_{1}\right)\right]\\
 & +\Expect_{\pi\left(a_{0}|s_{0}\right)}\left[Q_{\pi',\beta}\left(s_{0},a_{0}\right)-\beta\log\frac{\pi\left(a_{0}|s_{0}\right)}{\pi_{\text{ref}}\left(a_{0}|s_{0}\right)}-V_{\pi',\beta}\left(s_{0}\right)\right]\\
=\  & \gamma\Expect_{\pi\left(a_{0}|s_{0}\right)}\Expect_{p\left(s_{1}|s_{0},a_{0}\right)}\left[V_{\pi,\beta}\left(s_{1}\right)-V_{\pi',\beta}\left(s_{1}\right)\right]\\
 & +\Expect_{\pi\left(a_{0}|s_{0}\right)}\left[A_{\pi',\beta}\left(s_{0},a_{0}\right)-\beta\log\frac{\pi\left(a_{0}|s_{0}\right)}{\pi_{\text{ref}}\left(a_{0}|s_{0}\right)}\right]
\end{align*}

By expressing $V_{\pi}\left(s_{1}\right)-V_{\pi'}\left(s_{1}\right)$
in the same way as above, we have:
\begin{align*}
 & V_{\pi,\beta}\left(s_{0}\right)-V_{\pi',\beta}\left(s_{0}\right)\\
=\  & \gamma^{2}\Expect_{\pi\left(a_{0}|s_{0}\right)p\left(s_{1}|s_{0},a_{0}\right)\pi\left(a_{1}|s_{1}\right)p\left(s_{2}|s_{1},a_{1}\right)}\left[V_{\pi,\beta}\left(s_{2}\right)-V_{\pi',\beta}\left(s_{2}\right)\right]\\
 & +\gamma\Expect_{\pi\left(a_{0}|s_{0}\right)p\left(s_{1}|s_{0},a_{0}\right)\pi\left(a_{1}|s_{1}\right)}\left[A_{\pi',\beta}\left(s_{1},a_{1}\right)-\beta\log\frac{\pi\left(a_{1}|s_{1}\right)}{\pi_{\text{ref}}\left(a_{1}|s_{1}\right)}\right]\\
 & +\Expect_{\pi\left(a_{0}|s_{0}\right)}\left[A_{\pi',\beta}\left(s_{0},a_{0}\right)-\beta\log\frac{\pi\left(a_{0}|s_{0}\right)}{\pi_{\text{ref}}\left(a_{0}|s_{0}\right)}\right]
\end{align*}

By recursively applying this decomposition and noting that $\gamma^{t}\rightarrow0$
as $t\rightarrow\infty$, we have:
\begin{equation}
V_{\pi,\beta}\left(s_{0}\right)-V_{\pi',\beta}\left(s_{0}\right)=\Expect_{\tau\sim\pi\left(\cdot|s_{0}\right)}\left[\sum^{\infty}_{t=0}\gamma^{t}\left(A_{\pi',\beta}\left(s_{t},a_{t}\right)-\beta\log\frac{\pi\left(a_{t}|s_{t}\right)}{\pi_{\text{ref}}\left(a_{t}|s_{t}\right)}\right)\right]\label{eq:difference_in_value_func}
\end{equation}

Taking expectation over the initial state distribution $p\left(s_{0}\right)$,
we obtain:
\begin{equation}
\mathcal{D}_{\beta}\left(\pi,\pi'\right):=\mathcal{J}_{\beta}\left(\pi\right)-\mathcal{J}_{\beta}\left(\pi'\right)=\Expect_{p\left(s_{0}\right)}\left[V_{\pi,\beta}\left(s_{0}\right)-V_{\pi',\beta}\left(s_{0}\right)\right]
\end{equation}

Substituting the previous result in Eq.~\ref{eq:difference_in_value_func}
yields:
\begin{equation}
\boxed{\mathcal{D}_{\beta}\left(\pi,\pi'\right)=\Expect_{\tau\sim\pi}\left[\sum^{\infty}_{t=0}\gamma^{t}\left(A_{\pi',\beta}\left(s_{t},a_{t}\right)-\beta\log\frac{\pi\left(a_{t}|s_{t}\right)}{\pi_{\text{ref}}\left(a_{t}|s_{t}\right)}\right)\right]}\label{eq:perf_diff_KL_reg_RL}
\end{equation}

Substituting $\pi'$ with $\pi^{*}$ and using the identity $A_{*,\beta}\left(s,a\right)=\beta\log\frac{\pi^{*}(a|s)}{\pi_{\text{ref}}\left(a|s\right)}$,
we have:
\begin{align*}
\mathcal{D}_{\beta}\left(\pi,\pi^{*}\right) & :=\mathcal{J}_{\beta}\left(\pi\right)-\mathcal{J}_{\beta}\left(\pi^{*}\right)\\
 & =\Expect_{p\left(s_{0}\right)}\Expect_{\tau\sim\pi\left(\cdot|s_{0}\right)}\left[\sum^{\infty}_{t=0}\gamma^{t}\left(A_{\pi^{*},\beta}\left(s_{t},a_{t}\right)-\beta\log\frac{\pi\left(a_{t}|s_{t}\right)}{\pi_{\text{ref}}\left(a_{t}|s_{t}\right)}\right)\right]\\
 & =\Expect_{p\left(s_{0}\right)}\Expect_{\tau\sim\pi\left(\cdot|s_{0}\right)}\left[\sum^{\infty}_{t=0}\gamma^{t}\left(\beta\log\frac{\pi^{*}\left(a_{t}|s_{t}\right)}{\pi_{\text{ref}}\left(a_{t}|s_{t}\right)}-\beta\log\frac{\pi\left(a_{t}|s_{t}\right)}{\pi_{\text{ref}}\left(a_{t}|s_{t}\right)}\right)\right]
\end{align*}
which is equivalent to:
\begin{equation}
\boxed{\mathcal{D}_{\beta}\left(\pi,\pi^{*}\right)=\beta\Expect_{\tau\sim\pi}\left[\sum^{\infty}_{t=0}\gamma^{t}\left(\log\pi^{*}\left(a_{t}|s_{t}\right)-\log\pi\left(a_{t}|s_{t}\right)\right)\right]}
\end{equation}

Expressing the trajectory expectation in terms of the discounted state
visitation distribution $d_{\pi,\gamma}\left(s\right):=\left(1-\gamma\right)\sum^{\infty}_{t=0}\gamma^{t}p\left(s_{t}=s\right)$,
we obtain:
\begin{equation}
\boxed{\mathcal{D}_{\beta}\left(\pi,\pi^{*}\right)=-\frac{\beta}{1-\gamma}\Expect_{d_{\pi,\gamma}\left(s\right)}\left[D_{\text{KL}}\left(\pi\left(\cdot|s\right)\|\pi^{*}\left(\cdot|s\right)\right)\right]}
\end{equation}

Since the KL divergence is non-negative, it follows that $\mathcal{D}_{\beta}\left(\pi,\pi^{*}\right)\leq0$,
which reflects the optimality of $\pi^{*}$, i.e., $\mathcal{J}_{\beta}\left(\pi\right)\leq\mathcal{J}_{\beta}\left(\pi^{*}\right)$.
Equivalently, we can write:
\begin{equation}
\boxed{\mathcal{D}_{\beta}\left(\pi^{*},\pi\right)=-\mathcal{D}_{\beta}\left(\pi,\pi^{*}\right)=\frac{\beta}{1-\gamma}\Expect_{d_{\pi,\gamma}\left(s\right)}\left[D_{\text{KL}}\left(\pi\left(\cdot|s\right)\|\pi^{*}\left(\cdot|s\right)\right)\right]}
\end{equation}

Finally, in the trajectory-level formulation, this relation can be
expressed as:
\begin{equation}
\boxed{\mathcal{D}_{\beta}\left(\pi^{*},\pi\right)=\beta\Expect_{p\left(s_{0}\right)}\left[D_{\text{KL}}\left(\pi\left(\tau|s_{0}\right)\|\pi^{*}\left(\tau|s_{0}\right)\right)\right]}
\end{equation}

\subsection{Proof of Proposition~\ref{KL_policy_minimization}\label{subsec:Proof-of-KL-Policy-Grad}}

The gradient of the KL divergence $D_{\text{KL}}\left(\pi_{\theta}\left(\cdot|s\right)\|\pi'\left(\cdot|s\right)\right)$
is:

\begin{align}
 & \nabla_{\theta}D_{\text{KL}}\left(\pi_{\theta}\left(\cdot|s\right)\|\pi'\left(\cdot|s\right)\right)\nonumber \\
=\  & \nabla_{\theta}\int\pi_{\theta}\left(a|s\right)\log\frac{\pi_{\theta}\left(a|s\right)}{\pi'\left(a|s\right)}da\nonumber \\
=\  & \int\nabla_{\theta}\left(\pi_{\theta}\left(a|s\right)\log\frac{\pi_{\theta}\left(a|s\right)}{\pi'\left(a|s\right)}\right)da\nonumber \\
=\  & \int\log\frac{\pi_{\theta}\left(a|s\right)}{\pi'\left(a|s\right)}\nabla_{\theta}\pi_{\theta}\left(a|s\right)da+\int\pi_{\theta}\left(a|s\right)\nabla_{\theta}\log\frac{\pi_{\theta}\left(a|s\right)}{\pi'\left(a|s\right)}da\nonumber \\
=\  & \Expect_{\pi_{\theta}\left(a|s\right)}\left[\log\frac{\pi_{\theta}\left(a|s\right)}{\pi'\left(a|s\right)}\nabla_{\theta}\log\pi_{\theta}\left(a|s\right)\right]+\underbrace{\Expect_{\pi_{\theta}\left(a|s\right)}\left[\nabla_{\theta}\log\pi_{\theta}\left(a|s\right)\right]}_{=0}\label{eq:grad_KL_div}
\end{align}

The lass term equals zero because:
\begin{align*}
\Expect_{\pi_{\theta}\left(a|s\right)}\left[\nabla_{\theta}\log\pi_{\theta}\left(a|s\right)\right] & =\int\pi_{\theta}\left(a|s\right)\frac{\nabla_{\theta}\pi_{\theta}\left(a|s\right)}{\pi_{\theta}\left(a|s\right)}da\\
 & =\int\nabla_{\theta}\pi_{\theta}\left(a|s\right)da\\
 & =\nabla_{\theta}\int\pi_{\theta}\left(a|s\right)da\\
 & =\nabla_{\theta}1\\
 & =0
\end{align*}

Therefore, Eq.~\ref{eq:grad_KL_div} reduces to:
\[
\nabla_{\theta}D_{\text{KL}}\left(\pi_{\theta}\left(\cdot|s\right)\|\pi'\left(\cdot|s\right)\right)=\Expect_{\pi_{\theta}\left(\cdot|s\right)}\left[\log\frac{\pi_{\theta}\left(\cdot|s\right)}{\pi'\left(\cdot|s\right)}\nabla_{\theta}\log\pi_{\theta}\left(\cdot|s\right)\right]
\]

This gradient is similar to the gradient of the squared error loss:
\[
\nabla_{\theta}\Expect_{\pi_{\bar{\theta}}\left(\cdot|s\right)}\left[\left(\log\pi_{\theta}\left(\cdot|s\right)-\log\pi'\left(\cdot|s\right)\right)^{2}\right]=\Expect_{\pi_{\bar{\theta}}\left(\cdot|s\right)}\left[\log\frac{\pi_{\theta}\left(\cdot|s\right)}{\pi'\left(\cdot|s\right)}\nabla_{\theta}\log\pi_{\theta}\left(\cdot|s\right)\right]
\]

This implies that $D_{\text{KL}}\left(\pi_{\theta}\left(\cdot|s\right)\|\pi'\left(\cdot|s\right)\right)$
and $\Expect_{\pi_{\bar{\theta}}\left(\cdot|s\right)}\left[\left(\log\pi_{\theta}\left(\cdot|s\right)-\log\pi'\left(\cdot|s\right)\right)^{2}\right]$
are equivalent in terms of optimization.

\subsection{Gradient Analysis\label{subsec:Gradient-Analysis}}

Since policy gradient methods are only defined in the \emph{episodic}
setting, we restrict our analysis to this setting throughout.

\subsubsection{Gradient of the On-policy Advantage Matching loss}

The gradient of the on-policy Advantage Matching (AM) loss is given
by:
\begin{align*}
\nabla_{\theta}\mathcal{L}^{\text{on-policy}}_{\text{AM}}\left(\theta\right) & =\nabla_{\theta}\Expect_{\tau\sim\pi_{\bar{\theta}}}\left[\sum^{T-1}_{t=0}\frac{\gamma^{t}}{2\beta}\left(\beta\log\frac{\pi_{\theta}\left(a_{t}|s_{t}\right)}{\pi_{\text{ref}}\left(a_{t}|s_{t}\right)}-A_{\pi_{\bar{\theta}},\beta}\left(s_{t},a_{t}\right)\right)^{2}\right]\\
 & =\Expect_{\tau\sim\pi_{\bar{\theta}}}\left[\sum^{T-1}_{t=0}\frac{\gamma^{t}}{2\beta}\nabla_{\theta}\left(\beta\log\frac{\pi_{\theta}\left(a|s\right)}{\pi_{\text{ref}}\left(a|s\right)}-A_{\pi_{\bar{\theta}},\beta}\left(s,a\right)\right)^{2}\right]
\end{align*}

Applying the chain rule and simplifying yields:
\begin{equation}
\boxed{\nabla_{\theta}\mathcal{L}^{\text{on-policy}}_{\text{AM}}\left(\theta\right)=\Expect_{\tau\sim\pi_{\bar{\theta}}}\left[\sum^{T-1}_{t=0}\gamma^{t}\left(\beta\log\frac{\pi_{\theta}\left(a_{t}|s_{t}\right)}{\pi_{\text{ref}}\left(a_{t}|s_{t}\right)}-A_{\pi_{\bar{\theta}},\beta}\left(s_{t},a_{t}\right)\right)\nabla_{\theta}\log\pi_{\theta}\left(a_{t}|s_{t}\right)\right]}\label{eq:grad_onpolicy_AM}
\end{equation}

\subsubsection{Gradient of the KL-regularized RL objective}

The KL-regularized expected return $\mathcal{J_{\beta}}\left(\pi\right)$
can be decomposed into two components: a standard reward term and
a KL-regularization term:
\begin{align}
\mathcal{J}_{\beta}\left(\theta\right) & =\underbrace{\Expect_{\tau\sim\pi_{\theta}}\left[\sum^{T-1}_{t=0}\gamma^{t}r\left(s_{t},a_{t}\right)\right]}_{\text{reward term}}-\beta\underbrace{\Expect_{\tau\sim\pi_{\theta}}\left[\sum^{T-1}_{t=0}\gamma^{t}\left(\log\pi_{\theta}\left(a_{t}|s_{t}\right)-\log\pi_{\text{ref}}\left(a_{t}|s_{t}\right)\right)\right]}_{\text{KL-regularization term}}
\end{align}
where $\tau:=\left(s_{0},a_{0},s_{1},a_{1},...,s_{T-1},a_{T-1},s_{T}\right)$
denotes a trajectory sampled from $\pi$.

This can be written more compactly as:
\[
\mathcal{J}_{\beta}\left(\theta\right)=\Expect_{\tau\sim\pi_{\theta}}\left[\mathcal{R}\left(\tau\right)\right]-\beta\Expect_{\tau\sim\pi_{\theta}}\left[\mathcal{K}_{\pi_{\theta}}\left(\tau\right)\right]
\]
where $\mathcal{R}\left(\tau\right):=\sum^{T-1}_{t=0}\gamma^{t}r\left(s_{t},a_{t}\right)$
denotes the return along $\tau$, and $\mathcal{K}_{\pi_{\theta}}\left(\tau\right):=\sum^{T-1}_{t=0}\gamma^{t}\left(\log\pi_{\theta}\left(a_{t}|s_{t}\right)-\log\pi_{\text{ref}}\left(a_{t}|s_{t}\right)\right)$
denotes the cumulative discounted log ratio between the current policy
and the reference policy along $\tau$.

Taking gradients w.r.t. $\theta$, we obtain:
\begin{equation}
\nabla_{\theta}\mathcal{J}_{\beta}\left(\theta\right)=\nabla_{\theta}\Expect_{\tau\sim\pi_{\theta}}\left[\mathcal{R}\left(\tau\right)\right]-\beta\nabla_{\theta}\Expect_{\tau\sim\pi_{\theta}}\left[\mathcal{K}_{\pi_{\theta}}\left(\tau\right)\right]\label{eq:grad_expect_return}
\end{equation}

We now analyze the gradients of the two components separately.

\paragraph{Gradient of the reward term}

The gradient of the reward term can be estimated using the log-derivative
trick as:
\begin{align*}
\nabla_{\theta}\Expect_{\tau\sim\pi_{\theta}}\left[\mathcal{R}\left(\tau\right)\right] & =\nabla_{\theta}\int p_{\pi_{\theta}}\left(\tau\right)\mathcal{R}\left(\tau\right)d\tau\\
 & =\int\left(\nabla_{\theta}p_{\pi_{\theta}}\left(\tau\right)\right)\mathcal{R}\left(\tau\right)d\tau\\
 & =\int p_{\pi_{\theta}}\left(\tau\right)\left(\nabla_{\theta}\log p_{\pi_{\theta}}\left(\tau\right)\right)\mathcal{R}\left(\tau\right)d\tau
\end{align*}
which is equivalent to:
\begin{equation}
\boxed{\nabla_{\theta}\Expect_{\tau\sim\pi_{\theta}}\left[\mathcal{R}\left(\tau\right)\right]=\Expect_{\tau\sim\pi_{\theta}}\left[\mathcal{R}\left(\tau\right)\nabla_{\theta}\log p_{\pi_{\theta}}\left(\tau\right)\right]}\label{eq:grad_reward_term}
\end{equation}

\paragraph{Gradient of the KL-regularization term}

Unlike the return $\mathcal{R}\left(\tau\right)$, the term$\mathcal{K}_{\pi_{\theta}}\left(\tau\right)$
depends explicitly on $\theta$, leading to a different gradient decomposition:
\begin{align*}
\nabla_{\theta}\Expect_{\tau\sim\pi_{\theta}}\left[\mathcal{K}_{\pi_{\theta}}\left(\tau\right)\right] & =\nabla_{\theta}\int p_{\pi_{\theta}}\left(\tau\right)\mathcal{K}_{\pi_{\theta}}\left(\tau\right)d\tau\\
 & =\int\nabla_{\theta}\left(p_{\pi_{\theta}}\left(\tau\right)\mathcal{K}_{\pi_{\theta}}\left(\tau\right)\right)d\tau\\
 & =\int\left(p_{\pi_{\theta}}\left(\tau\right)\nabla_{\theta}\mathcal{K}_{\pi_{\theta}}\left(\tau\right)+\mathcal{K}_{\pi_{\theta}}\left(\tau\right)\nabla_{\theta}p_{\pi_{\theta}}\left(\tau\right)\right)d\tau
\end{align*}
which can be equivalently written as:
\[
\nabla_{\theta}\Expect_{\tau\sim\pi_{\theta}}\left[\mathcal{K}_{\pi_{\theta}}\left(\tau\right)\right]=\underbrace{\Expect_{\tau\sim\pi_{\theta}}\left[\nabla_{\theta}\mathcal{K}_{\pi_{\theta}}\left(\tau\right)\right]}_{=0}+\Expect_{\tau\sim\pi_{\theta}}\left[\mathcal{K}_{\pi_{\theta}}\left(\tau\right)\nabla_{\theta}\log p_{\pi_{\theta}}\left(\tau\right)\right]
\]

The gradient of $\mathcal{K}_{\pi_{\theta}}\left(\tau\right)$ is:
\begin{align*}
\nabla_{\theta}\mathcal{K}_{\pi_{\theta}}\left(\tau\right) & =\nabla_{\theta}\left(\sum^{T-1}_{t=0}\gamma^{t}\left(\log\pi_{\theta}\left(a_{t}|s_{t}\right)-\log\pi_{\text{ref}}\left(a_{t}|s_{t}\right)\right)\right)\\
 & =\sum^{T-1}_{t=0}\gamma^{t}\nabla_{\theta}\log\pi_{\theta}\left(a_{t}|s_{t}\right)
\end{align*}

Taking expectation, we have:
\begin{align*}
\Expect_{\tau\sim\pi_{\theta}}\left[\nabla_{\theta}\mathcal{K}_{\pi_{\theta}}\left(\tau\right)\right] & =\Expect_{\tau\sim\pi_{\theta}}\left[\sum^{T-1}_{t=0}\gamma^{t}\nabla_{\theta}\log\pi_{\theta}\left(a_{t}|s_{t}\right)\right]\\
 & =\frac{1}{1-\gamma}\Expect_{d_{\pi_{\theta},\gamma}\left(s\right)}\Expect_{\pi_{\theta}\left(a|s\right)}\left[\nabla_{\theta}\log\pi_{\theta}\left(a|s\right)\right]
\end{align*}

Using the identity $\Expect_{\pi_{\theta}\left(a|s\right)}\left[\nabla_{\theta}\log\pi_{\theta}\left(a|s\right)\right]=0$,
we obtain: 
\[
\Expect_{\tau\sim\pi_{\theta}}\left[\nabla_{\theta}\mathcal{K}_{\pi_{\theta}}\left(\tau\right)\right]=0
\]

Therefore, the gradient of the KL-regularization term reduces to:
\begin{equation}
\boxed{\nabla_{\theta}\Expect_{\tau\sim\pi_{\theta}}\left[\mathcal{K}_{\pi_{\theta}}\left(\tau\right)\right]=\Expect_{\tau\sim\pi_{\theta}}\left[\mathcal{K}_{\pi_{\theta}}\left(\tau\right)\nabla_{\theta}\log p_{\pi_{\theta}}\left(\tau\right)\right]}\label{eq:grad_KL_reg_term}
\end{equation}

\paragraph{Gradient of the KL-regularized RL Objective}

Substituting the gradients of the reward term (Eq.~\ref{eq:grad_reward_term})
and the KL-regularization term (Eq.~\ref{eq:grad_KL_reg_term}) into
Eq.~\ref{eq:grad_expect_return} yields:
\[
\boxed{\nabla_{\theta}\mathcal{J}_{\beta}\left(\theta\right)=\Expect_{\tau\sim\pi_{\theta}}\left[\left(\mathcal{R}\left(\tau\right)-\beta\mathcal{K}_{\pi_{\theta}}\left(\tau\right)\right)\nabla_{\theta}\log p_{\pi_{\theta}}\left(\tau\right)\right]}
\]

Define the KL-regularized return:
\[
\tilde{\mathcal{R}}\left(\tau\right):=\mathcal{R}\left(\tau\right)-\beta\mathcal{K}_{\pi_{\theta}}\left(\tau\right)=\sum^{T-1}_{t=0}\gamma^{t}\left(r\left(s_{t},a_{t}\right)-\beta\log\frac{\pi_{\theta}\left(a_{t}|s_{t}\right)}{\pi_{\text{ref}}\left(a_{t}|s_{t}\right)}\right)=\sum^{T-1}_{t=0}\gamma^{t}\tilde{r}\left(s_{t},a_{t}\right)
\]

Then $\nabla_{\theta}\mathcal{J}_{\beta}\left(\theta\right)$ can
be written compactly as:
\begin{equation}
\nabla_{\theta}\mathcal{J}_{\beta}\left(\theta\right)=\Expect_{\tau\sim\pi_{\theta}}\left[\tilde{\mathcal{R}}\left(\tau\right)\nabla_{\theta}\log p_{\pi_{\theta}}\left(\tau\right)\right]\label{eq:grad_KL_reg_RL_obj_R_tilde}
\end{equation}

Using the trajectory log-probability decomposition:
\[
\nabla_{\theta}\log p_{\pi_{\theta}}\left(\tau\right)=\sum^{T-1}_{t=0}\nabla_{\theta}\log\pi_{\theta}\left(a_{t}|s_{t}\right)
\]
we obtain:
\begin{equation}
\nabla_{\theta}\mathcal{J}_{\beta}\left(\theta\right)=\Expect_{\tau\sim\pi_{\theta}}\left[\tilde{\mathcal{R}}\left(\tau\right)\sum^{T-1}_{t=0}\nabla_{\theta}\log\pi_{\theta}\left(a_{t}|s_{t}\right)\right]\label{eq:grad_KL_reg_RL_obj_R_tilde_2}
\end{equation}

We can split $\nabla_{\theta}\mathcal{J}_{\beta}\left(\theta\right)$
into the past and future terms as follows:
\begin{align*}
 & \nabla_{\theta}\mathcal{J}_{\beta}\left(\theta\right)\\
=\  & \Expect_{\tau\sim\pi_{\theta}}\left[\sum^{T-1}_{t=0}\tilde{\mathcal{R}}\left(\tau\right)\nabla_{\theta}\log\pi_{\theta}\left(a_{t}|s_{t}\right)\right]\\
=\  & \Expect_{\tau\sim\pi_{\theta}}\left[\sum^{T-1}_{t=0}\left(\sum^{T-1}_{t'=0}\gamma^{t'}\tilde{r}\left(s_{t'},a_{t'}\right)\right)\nabla_{\theta}\log\pi_{\theta}\left(a_{t}|s_{t}\right)\right]\\
=\  & \underbrace{\Expect_{\tau\sim\pi_{\theta}}\left[\sum^{T-1}_{t=0}\sum_{t'<t}\gamma^{t'}\tilde{r}\left(s_{t'},a_{t'}\right)\nabla_{\theta}\log\pi_{\theta}\left(a_{t}|s_{t}\right)\right]}_{=0}+\Expect_{\tau\sim\pi_{\theta}}\left[\sum^{T-1}_{t=0}\sum_{t'\geq t}\gamma^{t'}\tilde{r}\left(s_{t'},a_{t'}\right)\nabla_{\theta}\log\pi_{\theta}\left(a_{t}|s_{t}\right)\right]
\end{align*}

The first term vanishes due to:
\begin{align*}
 & \Expect_{\tau\sim\pi_{\theta}}\left[\sum^{T-1}_{t=0}\sum_{t'<t}\gamma^{t'}\tilde{r}\left(s_{t'},a_{t'}\right)\nabla_{\theta}\log\pi_{\theta}\left(a_{t}|s_{t}\right)\right]\\
=\  & \sum^{T-1}_{t=0}\sum_{t'<t}\Expect_{\tau\sim\pi_{\theta}}\left[\gamma^{t'}\tilde{r}\left(s_{t'},a_{t'}\right)\nabla_{\theta}\log\pi_{\theta}\left(a_{t}|s_{t}\right)\right]\\
=\  & \sum^{T-1}_{t=0}\sum_{t'<t}\Expect_{p_{\pi_{\theta}}\left(s_{t}\right)}\Expect_{\pi_{\theta}\left(a_{t}|s_{t}\right)}\left[\gamma^{t'}\tilde{r}\left(s_{t'},a_{t'}\right)\nabla_{\theta}\log\pi_{\theta}\left(a_{t}|s_{t}\right)\right]\\
=\  & \sum^{T-1}_{t=0}\sum_{t'<t}\Expect_{p_{\pi_{\theta}}\left(s_{t}\right)}\left[\gamma^{t'}\tilde{r}\left(s_{t'},a_{t'}\right)\underbrace{\Expect_{\pi_{\theta}\left(a_{t}|s_{t}\right)}\left[\nabla_{\theta}\log\pi_{\theta}\left(a_{t}|s_{t}\right)\right]}_{=0}\right]\\
=\  & 0
\end{align*}

Consequently,
\[
\nabla_{\theta}\mathcal{J}_{\beta}\left(\theta\right)=\Expect_{\tau\sim\pi_{\theta}}\left[\sum^{T-1}_{t=0}\sum_{t'\geq t}\gamma^{t'}\tilde{r}\left(s_{t'},a_{t'}\right)\nabla_{\theta}\log\pi_{\theta}\left(a_{t}|s_{t}\right)\right]
\]

Using the law of iterated expectations, we have:
\begin{align*}
\nabla_{\theta}\mathcal{J}_{\beta}\left(\theta\right) & =\Expect_{\tau\sim\pi_{\theta}}\left[\sum^{T-1}_{t=0}\gamma^{t}\sum_{t'\geq t}\gamma^{\left(t'-t\right)}\tilde{r}\left(s_{t'},a_{t'}\right)\nabla_{\theta}\log\pi_{\theta}\left(a_{t}|s_{t}\right)\right]\\
 & =\Expect_{\tau\sim\pi_{\theta}}\left[\sum^{T-1}_{t=0}\gamma^{t}\left(\underbrace{r\left(s_{t},a_{t}\right)+\Expect_{\tau'\sim\pi_{\theta}}\left[\sum_{t'>t}\gamma^{\left(t'-t\right)}\tilde{r}\left(s_{t'},a_{t'}\right)\right]}_{Q_{\pi_{\theta},\beta}\left(s_{t},a_{t}\right)}-\beta\log\frac{\pi_{\theta}\left(a_{t}|s_{t}\right)}{\pi_{\text{ref}}\left(a_{t}|s_{t}\right)}\right)\nabla_{\theta}\log\pi_{\theta}\left(a_{t}|s_{t}\right)\right]\\
 & =\Expect_{\tau\sim\pi_{\theta}}\left[\sum^{T-1}_{t=0}\gamma^{t}\left(Q_{\pi_{\theta},\beta}\left(s_{t},a_{t}\right)-\beta\log\frac{\pi_{\theta}\left(a_{t}|s_{t}\right)}{\pi_{\text{ref}}\left(a_{t}|s_{t}\right)}\right)\nabla_{\theta}\log\pi_{\theta}\left(a_{t}|s_{t}\right)\right]
\end{align*}

Introducing $V_{\pi_{\theta},\beta}\left(s_{t}\right)$ as a baseline
yields:
\[
\nabla_{\theta}\mathcal{J}_{\beta}\left(\theta\right)=\Expect_{\tau\sim\pi_{\theta}}\left[\sum^{T-1}_{t=0}\gamma^{t}\left(Q_{\pi_{\theta},\beta}\left(s_{t},a_{t}\right)-V_{\pi_{\theta},\beta}\left(s_{t}\right)-\beta\log\frac{\pi_{\theta}\left(a_{t}|s_{t}\right)}{\pi_{\text{ref}}\left(a_{t}|s_{t}\right)}\right)\nabla_{\theta}\log\pi_{\theta}\left(a_{t}|s_{t}\right)\right]
\]
\[
=\Expect_{\tau\sim\pi_{\theta}}\left[\sum^{T-1}_{t=0}\gamma^{t}\left(A_{\pi_{\theta},\beta}\left(s_{t},a_{t}\right)-\beta\log\frac{\pi_{\theta}\left(a_{t}|s_{t}\right)}{\pi_{\text{ref}}\left(a_{t}|s_{t}\right)}\right)\nabla_{\theta}\log\pi_{\theta}\left(a_{t}|s_{t}\right)\right]
\]

Taking negative of both sides, we obtain:
\begin{equation}
\boxed{-\nabla_{\theta}\mathcal{J}_{\beta}\left(\theta\right)=\Expect_{\tau\sim\pi_{\theta}}\left[\sum^{T-1}_{t=0}\gamma^{t}\left(\beta\log\frac{\pi_{\theta}\left(a_{t}|s_{t}\right)}{\pi_{\text{ref}}\left(a_{t}|s_{t}\right)}-A_{\pi_{\theta},\beta}\left(s_{t},a_{t}\right)\right)\nabla_{\theta}\log\pi_{\theta}\left(a_{t}|s_{t}\right)\right]}\label{eq:grad_KL_reg_RL_obj_final}
\end{equation}

\subsubsection{Comparison between the gradients of the On-policy Advantage Matching
loss and the KL-regularized RL objective}

From Eq.~\ref{eq:grad_onpolicy_AM} and Eq.~\ref{eq:grad_KL_reg_RL_obj_final},
it follows that:
\[
\boxed{\nabla_{\theta}\mathcal{L}^{\text{on-policy}}_{\text{AM}}\left(\theta\right)=-\nabla_{\theta}\mathcal{J}_{\beta}\left(\theta\right)}
\]
That is, minimizing the on-policy Advantage Matching loss is equivalent
to maximizing the KL-regularized expected return.

\subsection{Different Off-Policy Variants\label{subsec:Different-Off-Policy-Variants}}

The trajectory-level soft advantage under the sampling policy $\pi_{\text{s}}$
can be estimated via Monte Carlo as:
\begin{align*}
A_{\pi_{\text{s}},\beta}\left(s_{0},\tau\right) & =Q_{\pi_{\text{s}},\beta}\left(s_{0},\tau\right)-\Expect_{\tau'\sim\pi_{\text{s}}\left(\cdot|s_{0}\right)}\left[Q_{\pi_{\text{s}},\beta}\left(s_{0},\tau'\right)-\beta\log\frac{\pi_{\text{s}}\left(\tau'|s_{0}\right)}{\pi_{\text{ref}}\left(\tau'|s_{0}\right)}\right]\\
 & \approx\mathcal{R}\left(\tau\right)-\Expect_{\tau'\sim\pi_{\text{s}}\left(\cdot|s_{0}\right)}\left[\mathcal{R}\left(\tau'\right)-\beta\log\frac{\pi_{\text{s}}\left(\tau'|s_{0}\right)}{\pi_{\text{ref}}\left(\tau'|s_{0}\right)}\right]\\
 & =\mathcal{R}\left(\tau\right)-B_{\pi_{\text{s}}}\left(s_{0}\right)
\end{align*}
where $B_{\pi_{\text{s}}}\left(s_{0}\right)$ denotes the corresponding
baseline.

A simple way to eliminate off-policy bias in advantage matching is
to discard the baseline $B_{\pi_{\text{s}}}\left(s\right)$ that depends
on $\pi_{\text{s}}$, yielding the \emph{trajectory-level Return Matching
(RM-Traj) loss}:
\begin{equation}
\Loss^{\text{off-policy}}_{\text{AM-Traj-v1}}\left(\theta\right)=\Expect_{p\left(s_{0}\right)}\Expect_{\tau\sim\pi_{\text{s}}\left(\cdot|s_{0}\right)}\left[\frac{1}{2\beta}w_{\bar{\theta}}\left(s_{0},\tau\right)\left(\beta\log\frac{\pi_{\theta}\left(\tau|s_{0}\right)}{\pi_{\text{ref}}\left(\tau|s_{0}\right)}-\mathcal{R}\left(\tau\right)\right)^{2}\right]\label{eq:RM_Traj_loss}
\end{equation}
where $w_{\bar{\theta}}\left(s_{0},\tau\right)=\frac{\pi_{\bar{\theta}}\left(\tau|s_{0}\right)}{\pi_{\text{s}}\left(\tau|s_{0}\right)}$
is the importance weight.

Notably, this objective can be interpreted as an \emph{online, positive-response-only}
variant of DPO \cite{rafailov2023direct}. Its main limitation, however,
is high variance. 

To reduce variance, instead of discarding $B_{\pi_{\text{s}}}\left(s_{0}\right)$
entirely, we use the baseline $B_{2}\left(s_{0}\right):=\Expect_{\tau'\sim\pi_{\text{s}}\left(\cdot|s_{0}\right)}\left[\mathcal{R}\left(\tau'\right)\right]$,
leading to the objective:
\begin{equation}
\Loss^{\text{off-policy}}_{\text{AM-Traj-v2}}\left(\theta\right)=\Expect_{p\left(s_{0}\right)}\Expect_{\tau\sim\pi_{\text{s}}\left(\cdot|s_{0}\right)}\left[\frac{1}{2\beta}w_{\bar{\theta}}\left(s_{0},\tau\right)\left(\beta\log\frac{\pi_{\theta}\left(\tau|s_{0}\right)}{\pi_{\text{ref}}\left(\tau|s_{0}\right)}-\left(\mathcal{R}\left(\tau\right)-\Expect_{\tau'\sim\pi_{\text{s}}\left(\cdot|s_{0}\right)}\left[\mathcal{R}\left(\tau'\right)\right]\right)\right)^{2}\right]\label{eq:HAM_Traj_loss}
\end{equation}

Since $\mathcal{R}\left(\tau\right)-\Expect_{\tau'\sim\pi_{\text{s}}\left(\cdot|s_{0}\right)}\left[\mathcal{R}\left(\tau'\right)\right]$
is the Monte Carlo estimator of the \emph{``hard''} trajectory-level
advantage $A_{\pi_{\text{s}}}\left(s_{0},\tau\right)\equiv A_{\pi_{\text{s}},0}\left(s_{0},\tau\right)$
in standard RL, we refer to this formulation as \emph{trajectory-level
Hard Advantage Matching (HAM-Traj)}.

Finally, when $\pi_{\text{s}}$ always stays close to $\pi_{\theta}$
during training (e.g., it is $\pi_{\theta}$ itself with different
sampling configurations or the last version of $\pi_{\theta}$), we
consider the objective:
\begin{equation}
\Loss^{\text{off-policy}}_{\text{AM-Traj-v3}}\left(\theta\right)=\Expect_{p\left(s_{0}\right)}\Expect_{\tau\sim\pi_{\text{s}}\left(\cdot|s_{0}\right)}\left[\frac{1}{2\beta}\left(\beta\log\left(\frac{\pi_{\theta}\left(\tau|s_{0}\right)}{\pi_{\text{ref}}\left(\tau|s_{0}\right)}\frac{\pi_{\text{s}}\left(\tau|s_{0}\right)}{\pi_{\bar{\theta}}\left(\tau|s_{0}\right)}\right)-A_{\pi_{\text{s}},\beta}\left(s_{0},\tau\right)\right)^{2}\right]\label{eq:AM4SP_Traj_loss}
\end{equation}

This formulation rescales the policy ratio $\frac{\pi_{\theta}\left(\tau|s_{0}\right)}{\pi_{\text{ref}}\left(\tau|s_{0}\right)}$
to yield a well-posed matching target of the form:
\[
\beta\log\frac{\pi_{\text{s}}\left(\tau|s_{0}\right)}{\pi_{\text{ref}}\left(\tau|s_{0}\right)}-A_{\pi_{\text{s}},\beta}\left(s_{0},\tau\right),
\]
thereby directly encouraging $\pi_{\text{s}}$ to approach optimality.
As $\pi_{\theta}$ is close to $\pi_{\text{s}}$, it also becomes
asymptotically optimal. However, significant deviation between $\pi_{\theta}$
and $\pi_{\text{s}}$ can introduce bias. We refer to this objective
as \emph{trajectory-level Advantage Matching for the Sampling Policy
(AM4SP-Traj)}.

\section{Additional Experimental Details}

\subsection{Benchmarks}

\subsubsection*{Math Reasoning}

Table~\ref{tab:Math-Benchmarks} presents the number of samples and
representative question-answer pairs for each math reasoning benchmark.

\begin{table}[H]
\begin{centering}
\resizebox{\textwidth}{!}{%
\begin{tabular}{cc>{\raggedright}p{0.65\textwidth}l}
\hline 
Benchmark & \#Samples & Question & Answer\tabularnewline
\hline 
\hline 
AIME24 & 30 & Define $f(x)=||x|-\tfrac{1}{2}|$ and $g(x)=||x|-\tfrac{1}{4}|$.
Find the number of intersections of the graphs of $y=4g(f(\sin(2\pi x)))$
and $x=4g(f(\cos(3\pi y)))$. & $\boxed{385}$\tabularnewline
AIME25 & 30 & Find the number of ordered pairs $(x,y)$, where both $x$ and $y$
are integers between $-100$ and $100$ inclusive, such that $12x^{2}-xy-6y^{2}=0$. & 117\tabularnewline
AMC23 & 40 & Positive real numbers x and y satisfy $y^{3}=x^{2}$ and $(y-x)^{2}=4y^{2}$.
What is $x+y$? & 36\tabularnewline
MATH-500 & 500 & If $f(x)=\frac{3x-2}{x-2}$, what is the value of $f(-2)+f(-1)+f(0)$?
Express your answer as a common fraction. & $\frac{14}{3}$\tabularnewline
Minerva Math & 272 & A particular star has an absolute magnitude $M=-7$. If this star
is observed in a galaxy that is at a distance of $3\mathrm{Mpc}$,
what will its apparent magnitude be? & 20.39\tabularnewline
OlympiadBench & 675 & Let$\mathbb{Q}_{>0}$ denote the set of all positive rational numbers.
Determine all functions $f:\mathbb{Q}_{>0}\rightarrow\mathbb{Q}_{>0}$
satisfying $f\left(x^{2}f(y)^{2}\right)=f(x)^{2}f(y)$ for all $x,y\in\mathbb{Q}_{>0}$. & $f(x)=1$\tabularnewline
\hline 
\end{tabular}}
\par\end{centering}
\caption{Overview of math reasoning benchmarks with representative examples.\label{tab:Math-Benchmarks}}
\end{table}

\subsubsection*{Code Generation}

Below, we report the number of samples and representative problem--test
case pairs for each code generation benchmark.

\paragraph{HumanEval}

164 samples

\textbf{\emph{Problem:}}

\begin{verbatim}
def has_close_elements(numbers: List[float], threshold: float) -> bool:
    """ Check if in given list of numbers, are any two numbers closer 
        to each other than given threshold.
    >>> has_close_elements([1.0, 2.0, 3.0], 0.5)
    False
    >>> has_close_elements([1.0, 2.8, 3.0, 4.0, 5.0, 2.0], 0.3)
    True
    """
\end{verbatim}

\textbf{\emph{Test:}}

\begin{verbatim}
def check(candidate):
    assert candidate([1.0, 2.0, 3.9, 4.0, 5.0, 2.2], 0.3) == True
    assert candidate([1.0, 2.0, 3.9, 4.0, 5.0, 2.2], 0.05) == False
    assert candidate([1.0, 2.0, 5.9, 4.0, 5.0], 0.95) == True
    assert candidate([1.0, 2.0, 5.9, 4.0, 5.0], 0.8) == False
    assert candidate([1.0, 2.0, 3.0, 4.0, 5.0, 2.0], 0.1) == True
    assert candidate([1.1, 2.2, 3.1, 4.1, 5.1], 1.0) == True
    assert candidate([1.1, 2.2, 3.1, 4.1, 5.1], 0.5) == False
\end{verbatim}

\paragraph{HumanEval+}

164 samples

\textbf{\emph{Problem:}}

\begin{verbatim}
def count_distinct_characters(string: str) -> int:
    """ Given a string, find out how many distinct characters 
        (regardless of case) does it consist of
    >>> count_distinct_characters('xyzXYZ')
    3
    >>> count_distinct_characters('Jerry')
    4
    """
\end{verbatim} 

\textbf{\emph{Test:}}

\begin{verbatim}
import numpy as np

def is_floats(x) -> bool:
    # check if it is float; List[float]; Tuple[float]
    if isinstance(x, float):
        return True
    if isinstance(x, (list, tuple)):
        return all(isinstance(i, float) for i in x)
    if isinstance(x, np.ndarray):
        return x.dtype == np.float64 or x.dtype == np.float32
    return False

def assertion(out, exp, atol):
    exact_match = out == exp

    if atol == 0 and is_floats(exp):
        atol = 1e-6
    if not exact_match and atol != 0:
        assert np.allclose(out, exp, rtol=1e-07, atol=atol)
    else:
        assert exact_match

def check(candidate):
    inputs = [[''], ['abcde'], ['abcdecadeCADE'], ['aaaaAAAAaaaa'],...]
    results = [0, 5, 5, 1,...]
    for i, (inp, exp) in enumerate(zip(inputs, results)):
        assertion(candidate(*inp), exp, 0)
\end{verbatim}

\paragraph{MBPP}

427 samples

\textbf{\emph{Problem:}} Write a python function to find the character
made by adding the ASCII value of all the characters of the given
string modulo 26. 

\textbf{\emph{Test:}} 

\begin{verbatim}
[assert get_Char("abc") == "f", 
 assert get_Char("gfg") == "t", 
 assert get_Char("ab") == "c"]
\end{verbatim}

\paragraph{MBPP+}

378 samples

\textbf{\emph{Problem:}} Write a python function to check whether
the given number can be represented as the difference of two squares
or not. 

\textbf{\emph{Test:}}

\begin{verbatim}
inputs = [[5], [10], [15], [991], [567],...]
results = [True, False, True, True, True,...]
for i, (inp, exp) in enumerate(zip(inputs, results)):
    assertion(dif_Square(*inp), exp, 0)
\end{verbatim}

\paragraph{LiveCodeBench}

877 samples

\textbf{\emph{Problem:}} You are given two integers \texttt{n} and
\texttt{x}. You have to construct an array of positive integers \texttt{nums}
of size \texttt{n} where for every $0\leq\texttt{i}\leq\texttt{n}-1$
, \texttt{nums[i+1]} is greater than \texttt{nums[i]}, and the result
of the bitwise AND operation between all elements of \texttt{nums}
is \texttt{x}. Return the minimum possible value of \texttt{nums[n-1]}.
Example 1: Input: \texttt{n = 3, x = 4} - Output: 6. Explanation:
\texttt{nums} can be \texttt{[4, 5, 6]} and its last element is
6. Example 2: Input: \texttt{n = 2, x = 7} - Output: 15. Explanation:
\texttt{nums} can be \texttt{[7, 15]} and its last element is 15.
Constraints: $1\leq\texttt{n}$ and $\texttt{x}\leq10^{8}$. 

\textbf{\emph{Test:}} 

\begin{verbatim}
[{"input": "3\\n4", "output": "6", "testtype": "functional"},
 {"input": "2\\n7", "output": "15", "testtype": "functional"}]
\end{verbatim}

\subsection{Prompt Templates}

Table~\ref{tab:Prompt_template} presents the prompt template used
throughout our experiments, which is consistently applied across all
models.

\begin{table}[H]
\begin{tabular}{>{\raggedright}p{0.99\textwidth}}
\hline 
\textbf{System (Math):} You are a mathematical solver assistant. You
first think about the solution process step by step and then provide
the user with the final answer.

When given a math problem:

1. Your response must strictly follow this structure: <think>Your
reasoning process</think> <answer> Your concluded solution and result
(use \textbackslash boxed\{\} to express the final result) </answer>.

2. The \textbackslash boxed\{\} expression must appear only once
in your conclusion so user can recognize the final result.

Result formats:

- yes/no question: The final answer should be ``Yes'' or ``No''

- true/false question: The final answer should be ``True'' or ``False''

- multiple choice questions: The final answer should be the choice
name only. For example: ``A'', ``B'', ``C'',...

- For other question: The final answer can be latex expressions (fractions,
polynomials, intervals, math symbols, equations...) or number.\tabularnewline
\tabularnewline
\textbf{System (Code): }You are a coding assistant. You first think
about the solution process step by step and then provides the user
with the final answer.

When given a coding problem:

1. Think (use <think> and </think> tags to include the thinking process)
then give step-by-step solution and final answer for the problem.

2. The ``\textasciigrave{}\textasciigrave{}\textasciigrave{}python
...\textasciigrave{}\textasciigrave{}\textasciigrave{}'' expression
must appear only once so user can recognize the final result

Result formats:

- The final answer should be a function implemented in Python. The
function name must be ``solution'' no matter what the original function
name is in the problem description.

- The function always returns a value. If the original function description
only mentions the function without a return statement or printing
out the result, you must modify it to return the result.\tabularnewline
\tabularnewline
\textbf{User:} Now provide solution for this problem: ``\textcolor{blue}{{[}question{]}}''\tabularnewline
\tabularnewline
\textbf{Assistant:} Let me solve this step by step.

<think>\tabularnewline
\hline 
\end{tabular}

\caption{The prompt template used in our experiments.\label{tab:Prompt_template}}
\end{table}

\section{Additional Experiment Results}

\subsection{Pass@$k$ Results\label{subsec:PassAtK-Results}}

\begin{table}
\begin{centering}
\resizebox{\textwidth}{!}{%
\begin{tabular}{cccccccc}
\hline 
\multicolumn{1}{c}{Method} & AIME24 & AIME25 & AMC23 & MATH-500 & Minerva & Olympiad & Avg.\tabularnewline
\hline 
\hline 
 & \multicolumn{7}{c}{Pass@2}\tabularnewline
GRPO & 9.3 & \textbf{5.0} & 37.0 & 58.3 & 31.4 & 22.4 & 27.2\tabularnewline
GSPO & 8.7 & 3.6 & 32.7 & 53.4 & 28.0 & 20.6 & 24.5\tabularnewline
Dr. GRPO & 10.6 & 3.2 & \textbf{41.9} & \multicolumn{1}{c}{58.8} & \textbf{33.3} & 24.2 & 28.7\tabularnewline
AM (ours) & \uline{11.5} & \uline{4.4} & 38.6 & \textbf{63.0} & \uline{33.1} & \uline{26.3} & \uline{29.5}\tabularnewline
AM-hybrid (ours) & \textbf{13.1} & \uline{4.4} & \uline{41.4} & \uline{62.9} & 32.8 & \textbf{27.0} & \textbf{30.3}\tabularnewline
\hline 
\hline 
 & \multicolumn{7}{c}{Pass@4}\tabularnewline
GRPO & 14.5 & \uline{7.9} & 46.9 & 67.4 & 37.4 & 30.3 & 34.1\tabularnewline
GSPO & 13.6 & 5.8 & 43.8 & 64.4 & 34.1 & 28.2 & 31.7\tabularnewline
Dr. GRPO & \uline{16.9} & 5.9 & \textbf{54.2} & 67.7 & \textbf{40.5} & 31.3 & 36.1\tabularnewline
AM (ours) & 16.1 & 6.9 & 48.7 & \textbf{71.4} & \uline{40.4} & \uline{33.3} & \uline{36.1}\tabularnewline
AM-hybrid (ours) & \textbf{19.2} & \textbf{8.0} & \uline{52.4} & \uline{71.3} & 39.6 & \textbf{34.0} & \textbf{37.4}\tabularnewline
\hline 
\hline 
 & \multicolumn{7}{c}{Pass@8}\tabularnewline
GRPO & 21.5 & \uline{11.8} & 57.8 & 74.9 & 43.2 & 38.5 & 41.3\tabularnewline
GSPO & 18.7 & 10.0 & 53.1 & 74.3 & 39.8 & 36.3 & 38.7\tabularnewline
Dr. GRPO & \textbf{24.6} & 9.8 & \textbf{64.9} & 75.6 & \uline{46.6} & 38.0 & 43.3\tabularnewline
AM (ours) & \uline{23.8} & 10.9 & 60.6 & \textbf{78.4} & \textbf{47.0} & \uline{40.1} & \uline{43.5}\tabularnewline
AM-hybrid (ours) & \textbf{24.6} & \textbf{14.3} & \uline{62.5} & \uline{77.9} & 45.6 & \textbf{41.1} & \textbf{44.3}\tabularnewline
\hline 
\end{tabular}}
\par\end{centering}
\caption{Pass@$k$ ($k\in\left\{ 2,4,8\right\} $) performance of Llama3.1-8B-Instruct
finetuned using our method and baseline online RL approaches across
multiple mathematical benchmarks. The best and second best results
are highlighted in \textbf{bold} and \uline{underlined}, respectively.\label{tab:PassK-table-Llama}}
\end{table}

\begin{table}
\begin{centering}
\resizebox{\textwidth}{!}{%
\begin{tabular}{cccccccc}
\hline 
\multicolumn{1}{c}{Method} & AIME24 & AIME25 & AMC23 & MATH-500 & Minerva & Olympiad & Avg.\tabularnewline
\hline 
\hline 
 & \multicolumn{7}{c}{Pass@2}\tabularnewline
GRPO & 26.5 & 16.1 & 69.8 & 84.9 & 39.6 & 46.1 & 47.2\tabularnewline
GSPO & 28.1 & 18.0 & 69.3 & 84.5 & 39.9 & 46.5 & 47.7\tabularnewline
Dr. GRPO & \uline{28.8} & 17.2 & 71.1 & \multicolumn{1}{c}{85.0} & 39.8 & 47.6 & 48.3\tabularnewline
AM (ours) & \textbf{30.4} & \uline{21.7} & \uline{72.7} & \uline{85.8} & \uline{40.0} & \uline{48.0} & \uline{49.8}\tabularnewline
AM-hybrid (ours) & 27.6 & \textbf{23.4} & \textbf{77.4} & \textbf{87.7} & \textbf{43.3} & \textbf{52.3} & \textbf{52.0}\tabularnewline
\hline 
\hline 
 & \multicolumn{7}{c}{Pass@4}\tabularnewline
GRPO & 35.4 & 22.3 & 78.7 & 88.6 & 44.4 & 52.3 & 53.6\tabularnewline
GSPO & 37.1 & 25.2 & 78.4 & 88.6 & \uline{45.2} & 53.8 & 54.7\tabularnewline
Dr. GRPO & \uline{38.7} & 24.4 & 79.3 & 88.6 & 44.8 & 54.2 & 55.0\tabularnewline
AM (ours) & \textbf{38.8} & \uline{27.7} & \uline{79.8} & \uline{89.4} & 44.4 & \uline{55.2} & \uline{55.9}\tabularnewline
AM-hybrid (ours) & 35.5 & \textbf{30.9} & \textbf{85.9} & \textbf{90.5} & \textbf{48.1} & \textbf{59.1} & \textbf{58.3}\tabularnewline
\hline 
\hline 
 & \multicolumn{7}{c}{Pass@8}\tabularnewline
GRPO & 43.4 & 31.1 & \uline{86.6} & 91.2 & 49.5 & 59.2 & 60.2\tabularnewline
GSPO & 45.9 & 31.7 & \uline{86.6} & 91.4 & 49.6 & 60.4 & 60.9\tabularnewline
Dr. GRPO & \textbf{49.3} & 32.4 & 86.4 & 91.5 & \uline{50.2} & 60.5 & \uline{61.7}\tabularnewline
AM (ours) & \uline{47.0} & \uline{34.6} & 85.2 & \uline{92.2} & 49.8 & \uline{61.4} & \uline{61.7}\tabularnewline
AM-hybrid (ours) & 42.9 & \textbf{38.4} & \textbf{92.4} & \textbf{92.7} & \textbf{52.8} & \textbf{65.3} & \textbf{64.1}\tabularnewline
\hline 
\end{tabular}}
\par\end{centering}
\caption{Pass@$k$ ($k\in\left\{ 2,4,8\right\} $) performance of Qwen2.5-Math-7B
finetuned using our method and baseline online RL approaches across
multiple mathematical benchmarks. The best and second best results
are highlighted in \textbf{bold} and \uline{underlined}, respectively.\label{tab:PassK-table-Qwen-Math}}
\end{table}

\begin{table}
\begin{centering}
\resizebox{\textwidth}{!}{%
\begin{tabular}{ccccccc}
\hline 
\multicolumn{1}{c}{Method} & HumanEval & HumanEval+ & MBPP & MBPP+ & LiveCodeBench & Avg.\tabularnewline
\hline 
\hline 
 & \multicolumn{6}{c}{Pass@2}\tabularnewline
GRPO & 86.7 & 81.1 & 79.4 & \uline{75.0} & 47.4 & 73.9\tabularnewline
GSPO & 85.6 & 79.9 & 79.1 & 74.8 & 47.7 & 73.4\tabularnewline
Dr. GRPO & 86.7 & 81.3 & \uline{80.2} & \multicolumn{1}{c}{\uline{75.0}} & 47.9 & 74.2\tabularnewline
AM (ours) & \uline{88.6} & \uline{84.0} & 79.2 & \textbf{75.6} & \textbf{49.0} & \uline{75.3}\tabularnewline
AM-hybrid (ours) & \textbf{90.8} & \textbf{85.5} & \textbf{80.3} & \uline{75.0} & \uline{48.2} & \textbf{76.0}\tabularnewline
\hline 
\hline 
 & \multicolumn{6}{c}{Pass@4}\tabularnewline
GRPO & 90.6 & 85.6 & 82.7 & \uline{78.1} & 54.6 & 78.3\tabularnewline
GSPO & 90.1 & 84.9 & 82.6 & 77.8 & 54.4 & 78.0\tabularnewline
Dr. GRPO & 90.8 & 85.6 & \uline{83.3} & 78.0 & \uline{54.9} & 78.5\tabularnewline
AM (ours) & \uline{92.1} & \uline{88.5} & 82.4 & \textbf{78.7} & \textbf{55.6} & \uline{79.5}\tabularnewline
AM-hybrid (ours) & \textbf{93.6} & \textbf{89.2} & \textbf{83.9} & \textbf{78.7} & \uline{54.9} & \textbf{80.1}\tabularnewline
\hline 
\hline 
 & \multicolumn{6}{c}{Pass@8}\tabularnewline
GRPO & 93.3 & 89.6 & \uline{85.5} & \uline{80.7} & 60.7 & 82.0\tabularnewline
GSPO & 93.2 & 88.6 & 85.2 & 80.1 & 60.2 & 81.5\tabularnewline
Dr. GRPO & 93.4 & 88.3 & 85.4 & 80.6 & \uline{60.8} & 81.7\tabularnewline
AM (ours) & \uline{94.8} & \uline{91.1} & 85.0 & \uline{80.7} & \textbf{61.1} & \uline{82.5}\tabularnewline
AM-hybrid (ours) & \textbf{95.3} & \textbf{91.4} & \textbf{86.5} & \textbf{81.6} & \uline{60.8} & \textbf{83.1}\tabularnewline
\hline 
\end{tabular}}
\par\end{centering}
\caption{Pass@$k$ ($k\in\left\{ 2,4,8\right\} $) performance of Qwen2.5-7B-Instruct
finetuned using our method and baseline online RL approaches across
multiple coding benchmarks. The best and second best results are highlighted
in \textbf{bold} and \uline{underlined}, respectively. \label{tab:PassK-Code-table-Qwen-Instr}}
\end{table}

\begin{table}
\begin{centering}
\resizebox{\textwidth}{!}{%
\begin{tabular}{ccccccc}
\hline 
\multicolumn{1}{c}{Method} & HumanEval & HumanEval+ & MBPP & MBPP+ & LiveCodeBench & Avg.\tabularnewline
\hline 
\hline 
 & \multicolumn{6}{c}{Pass@2}\tabularnewline
GRPO & 91.3 & \textbf{86.7} & \textbf{81.6} & \uline{76.3} & 50.1 & 77.2\tabularnewline
GSPO & 90.9 & 85.7 & 80.9 & 75.4 & 50.4 & 76.7\tabularnewline
Dr. GRPO & \uline{91.6} & 85.6 & 80.9 & \multicolumn{1}{c}{75.7} & 50.6 & 76.9\tabularnewline
AM (ours) & 91.4 & 86.4 & \textbf{81.6} & \textbf{76.8} & \uline{52.3} & \textbf{77.7}\tabularnewline
AM-hybrid (ours) & \textbf{92.0} & \uline{86.6} & \uline{81.1} & 75.6 & \textbf{53.2} & \textbf{77.7}\tabularnewline
\hline 
\hline 
 & \multicolumn{6}{c}{Pass@4}\tabularnewline
GRPO & 93.0 & 88.6 & \textbf{84.6} & \uline{79.0} & 56.6 & 80.4\tabularnewline
GSPO & \uline{93.4} & 88.0 & 84.2 & 78.4 & 56.5 & 80.1\tabularnewline
Dr. GRPO & \uline{93.4} & 88.0 & 84.2 & 78.8 & 56.6 & 80.2\tabularnewline
AM (ours) & 93.3 & \uline{88.5} & \uline{84.3} & \textbf{79.3} & \uline{58.1} & \uline{80.7}\tabularnewline
AM-hybrid (ours) & \textbf{94.3} & \textbf{89.5} & 83.8 & 78.6 & \textbf{59.2} & \textbf{81.1}\tabularnewline
\hline 
\hline 
 & \multicolumn{6}{c}{Pass@8}\tabularnewline
GRPO & \uline{94.6} & \uline{90.2} & \uline{86.5} & 81.1 & 62.1 & 82.9\tabularnewline
GSPO & \uline{94.6} & 90.0 & 86.4 & 81.1 & 61.6 & 82.7\tabularnewline
Dr. GRPO & 95.4 & 90.0 & \textbf{86.7} & 81.1 & 61.4 & 82.9\tabularnewline
AM (ours) & 94.3 & \uline{90.2} & \uline{86.5} & \uline{81.3} & \uline{63.4} & \uline{83.1}\tabularnewline
AM-hybrid (ours) & \textbf{95.8} & \textbf{91.8} & 86.0 & \textbf{81.9} & \textbf{64.5} & \textbf{84.0}\tabularnewline
\hline 
\end{tabular}}
\par\end{centering}
\caption{Pass@$k$ ($k\in\left\{ 2,4,8\right\} $) performance of Qwen2.5-Coder-7B-Instruct
finetuned using our method and baseline online RL approaches across
multiple coding benchmarks. The best and second best results are highlighted
in \textbf{bold} and \uline{underlined}, respectively. \label{tab:PassK-Code-table-Qwen-Coder}}
\end{table}

Tables \ref{tab:PassK-table-Llama} and \ref{tab:PassK-table-Qwen-Math}
report Pass@$k$ ($k\in\{2,4,8\}$) results for mathematical reasoning
across Llama-3.1-8B-Instruct and Qwen2.5-Math-7B, respectively. Similarly,
Tables \ref{tab:PassK-Code-table-Qwen-Instr} and \ref{tab:PassK-Code-table-Qwen-Coder}
present code generation results using Qwen2.5-7B-Instruct and Qwen2.5-Coder-7B-Instruct,
respectively. For mathematical reasoning, AM generally achieves higher
Pass@$k$ than competing baselines with Llama-3.1-8B-Instruct and
Qwen2.5-Math-7B. Notably, the improvements are more consistent with
math-specialized models, suggesting that AM effectively exploits their
intrinsic reasoning capabilities. For code generation, both AM and
AM-hybrid consistently outperform the baselines across the two backbones,
demonstrating that the benefits of our approach extend beyond mathematical
reasoning and generalize to code generation across both general-purpose
and code-specialized models.

\begin{figure}
\begin{centering}
\includegraphics[width=1\textwidth]{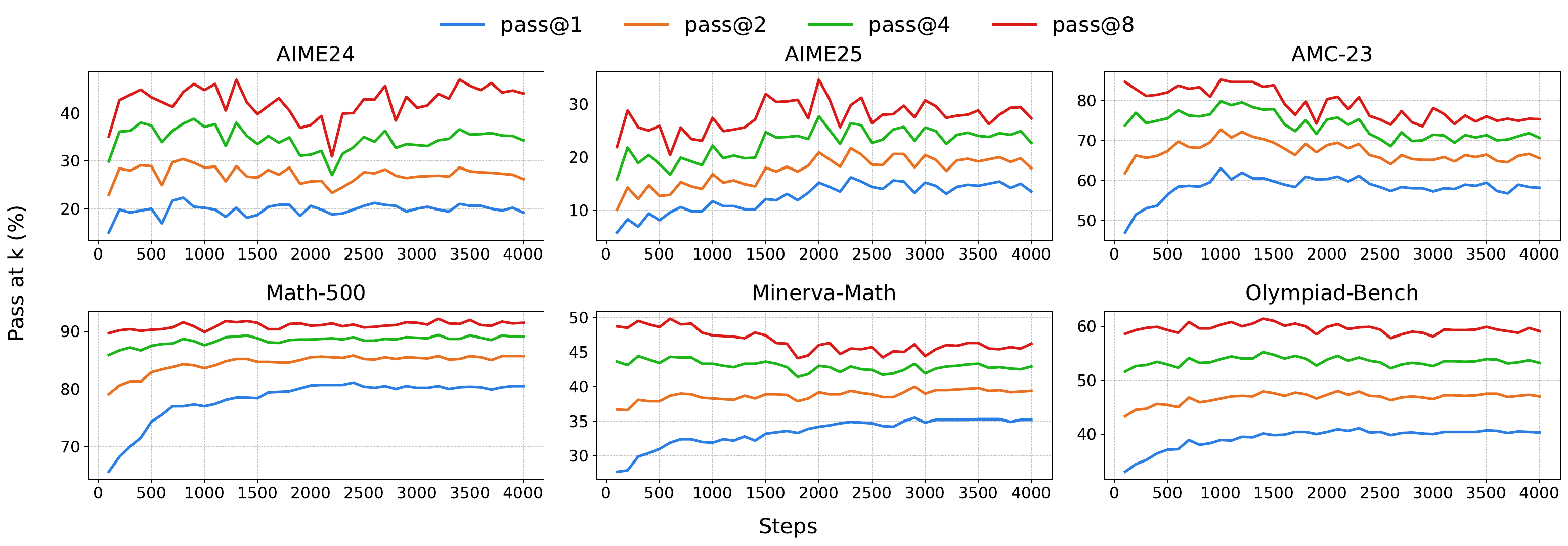}
\par\end{centering}
\caption{Pass@$k$ performance curves for AM across training iterations on
multiple benchmarks with $k\in\{1,2,4,8\}$. The base model is Qwen2.5-Math-7B.\label{fig:PassK-curves-Qwen-Math}}
\end{figure}

Figure~\ref{fig:PassK-curves-Qwen-Math} further visualizes the Pass@$k$
performance curves for Qwen2.5-Math-7B throughout training. Interestingly,
on certain benchmarks such as AMC-23 and Minerva Math, Pass@4 and
Pass@8 decrease as training progresses, suggesting reduced diversity
in generated solutions for more challenging problems in these datasets.
We attribute this behavior to the mode-seeking nature of RL optimization,
and observe that it consistently appears across all methods rather
than being specific to our approach.

\subsection{Additional Performance Curves\label{subsec:Additional-Performance-Curves}}

\begin{figure}
\centering{}\includegraphics[width=1\textwidth]{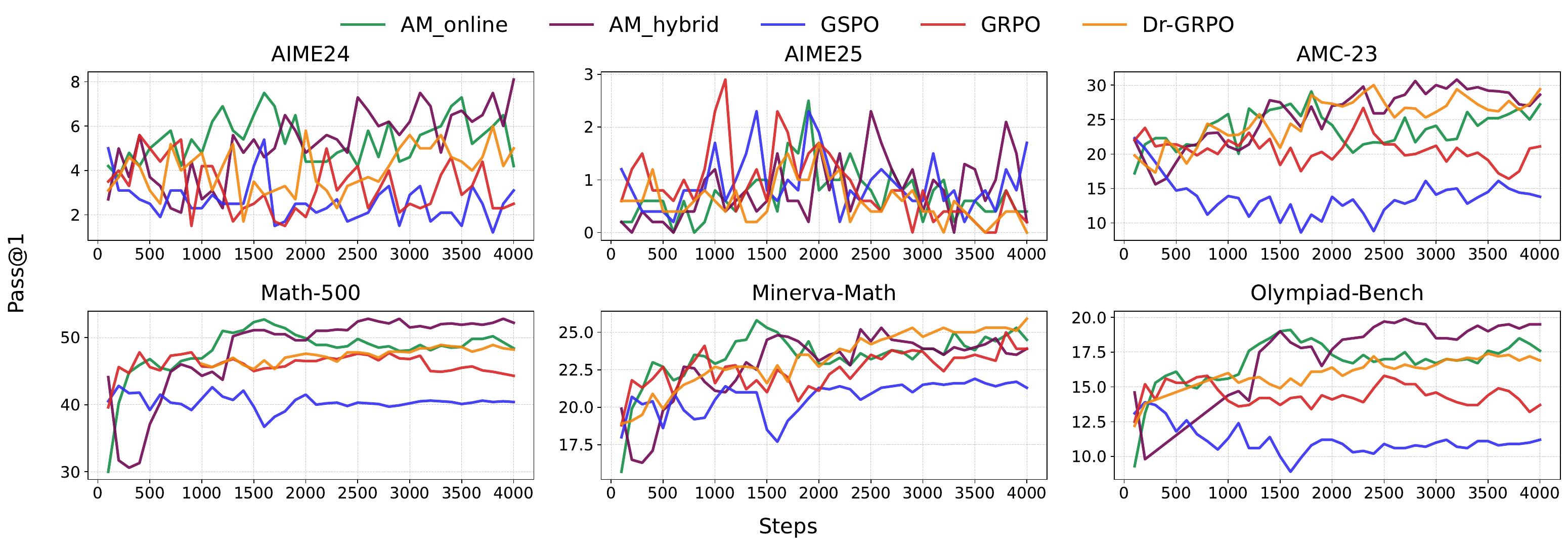}\caption{Pass@1 performance curves of Llama3.1-8B-Instruct finetuned with our
method and other RL baselines across multiple mathematical reasoning
benchmarks over the course of training.\label{fig:Pass-1-Math-curves-Llama}}
\end{figure}

\begin{figure}
\centering{}\includegraphics[width=1\textwidth]{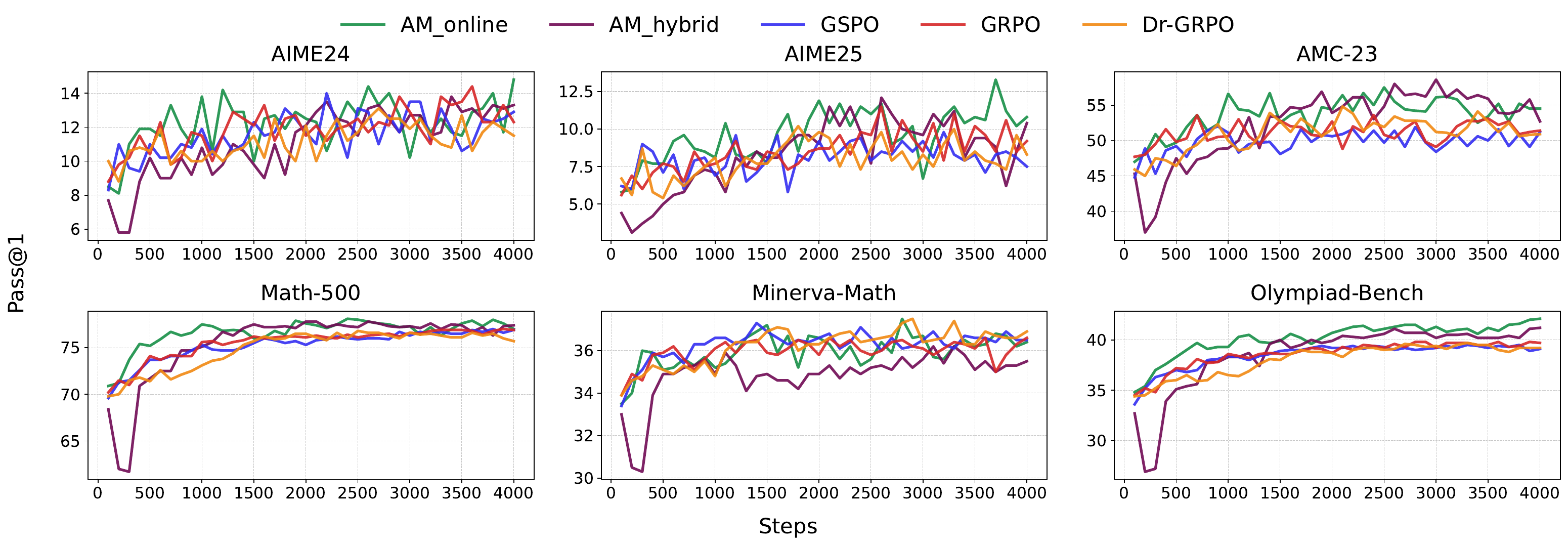}\caption{Pass@1 performance curves of Qwen2.5-7B-Instruct finetuned with our
method and other RL baselines across multiple mathematical reasoning
benchmarks over the course of training.\label{fig:Pass-1-Math-curves-Qwen-Instr}}
\end{figure}

\begin{figure}
\begin{centering}
\includegraphics[width=1\textwidth]{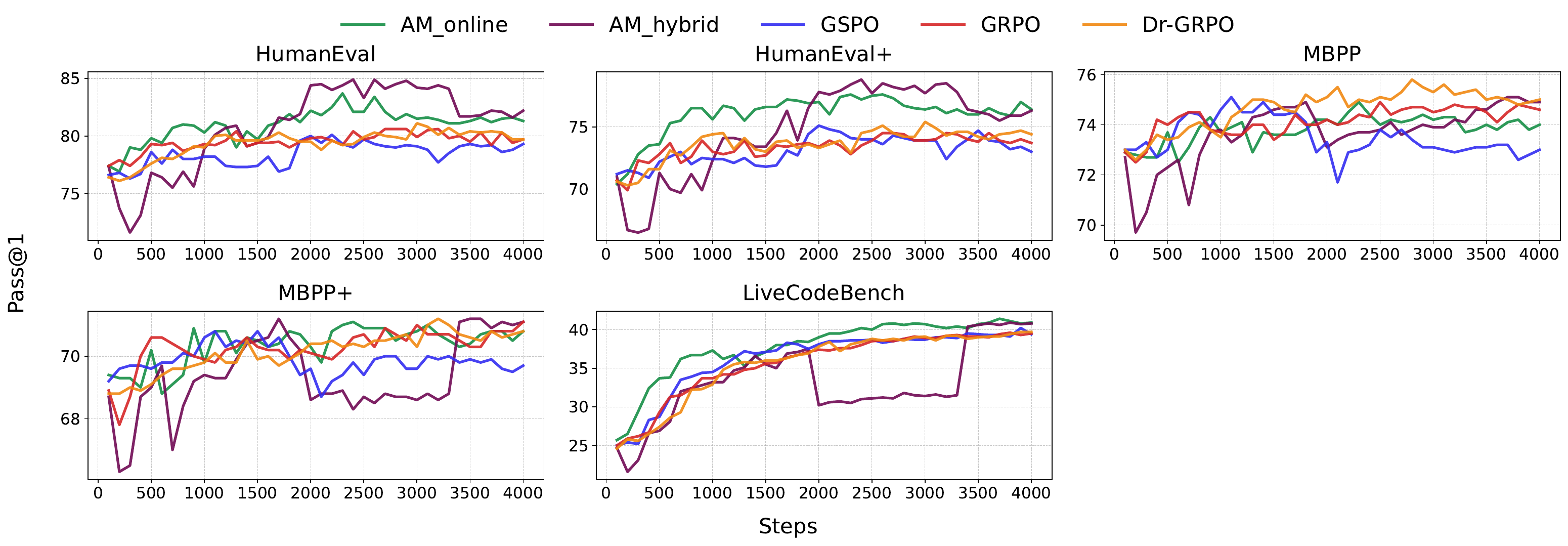}
\par\end{centering}
\caption{Pass@1 performance curves of Qwen2.5-7B-Instruct finetuned with our
method and other RL baselines across multiple code generation benchmarks
over the course of training. \label{fig:Pass-1-Code-curves-Qwen-Instr}}
\end{figure}

\begin{figure}
\begin{centering}
\includegraphics[width=1\textwidth]{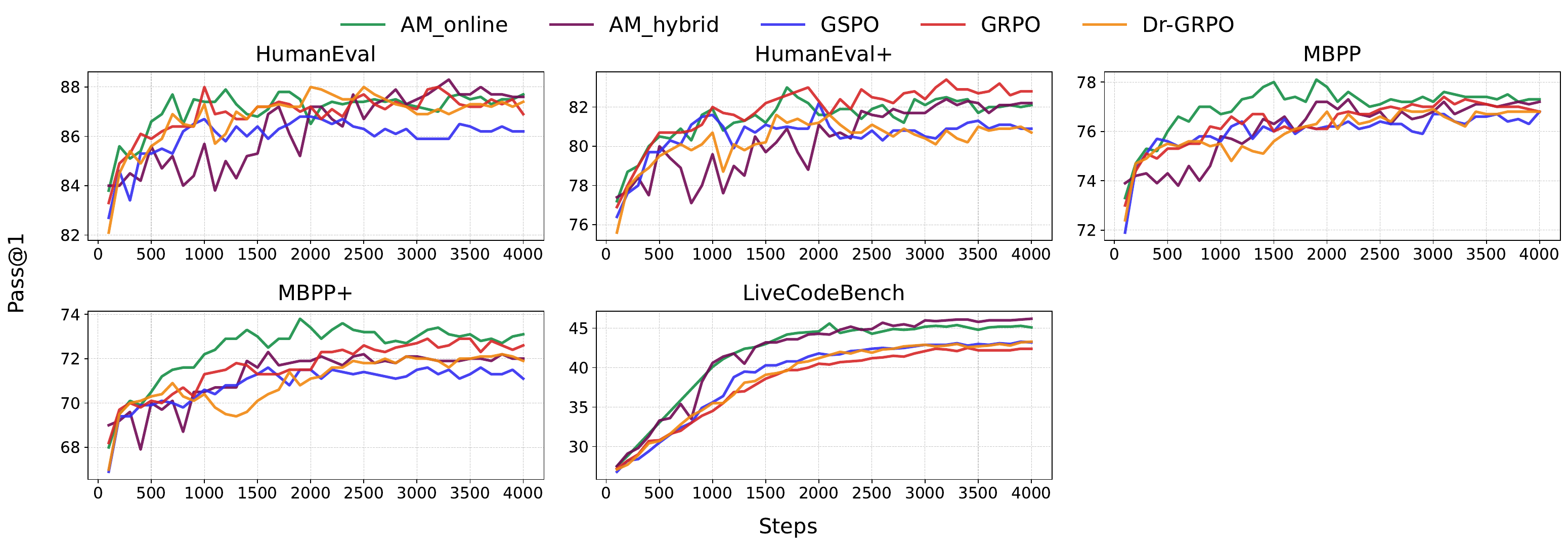}
\par\end{centering}
\caption{Pass@1 performance curves of Qwen2.5-Coder-7B-Instruct finetuned with
our method and other RL baselines across multiple code generation
benchmarks over the course of training. \label{fig:Pass-1-Code-curves-Qwen-Coder}}
\end{figure}

Figures~\ref{fig:Pass-1-Math-curves-Llama} and \ref{fig:Pass-1-Math-curves-Qwen-Instr}
illustrate the Pass@1 performance curves for mathematical reasoning
when using Llama3.1-8B-Instruct and Qwen2.5-7B-Instruct as the base
models, respectively. Meanwhile, \ref{fig:Pass-1-Code-curves-Qwen-Coder}
and \ref{fig:Pass-1-Code-curves-Qwen-Instr} show the Pass@1 performance
curves for code generation when finetuning Qwen2.5-Coder-7B-Instruct
and Qwen2.5-7B-Instruct.

\subsection{Trajectory Length Dynamics During Training\label{subsec:Trajectory-Length-Dynamics}}

\begin{figure}
\begin{centering}
\begin{tabular}{c}
\includegraphics[width=0.99\textwidth]{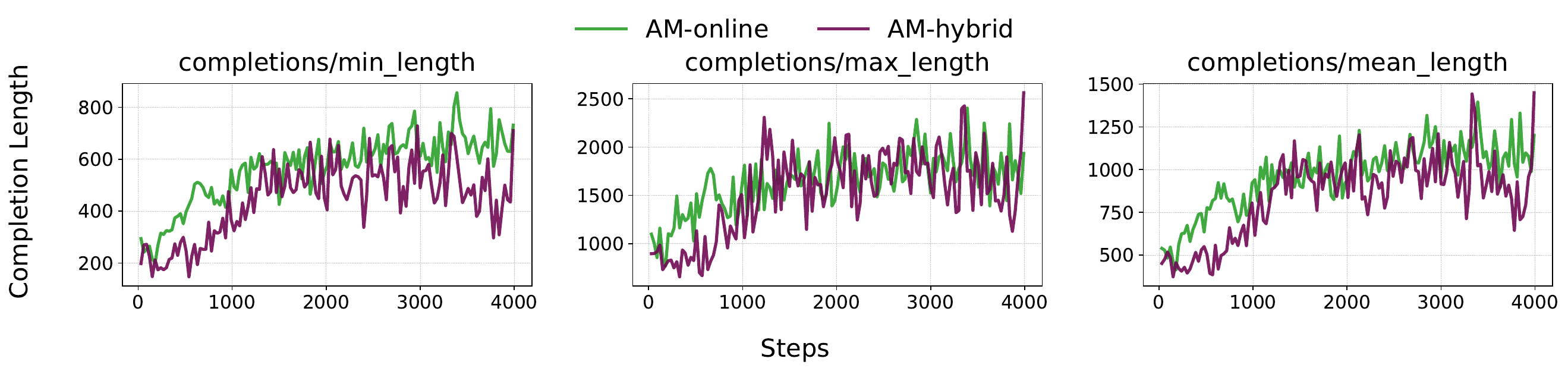}\tabularnewline
(a) Qwen2.5-7B-Instruct\tabularnewline
\tabularnewline
\includegraphics[width=0.99\textwidth]{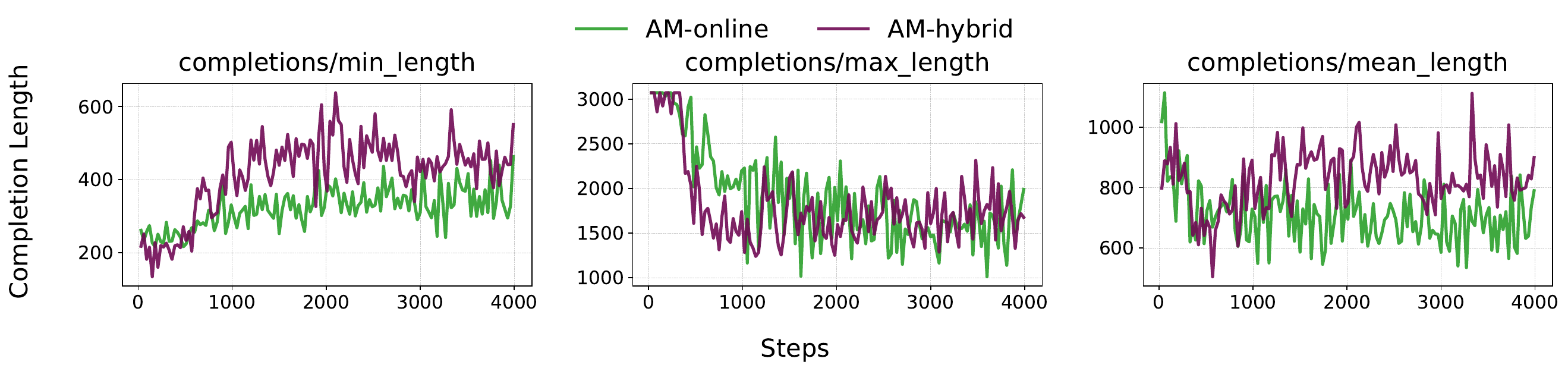}\tabularnewline
(b) Qwen2.5-Math-7B\tabularnewline
\end{tabular}
\par\end{centering}
\caption{Minimum, mean, and maximum completion lengths generated by AM over
the course of training.\label{fig:completion_length}}
\end{figure}

Figure~\ref{fig:completion_length} illustrates the evolution of
the maximum, mean, and minimum trajectory lengths generated by our
Advantage Matching (AM) method for Qwen2.5-7B-Instruct and Qwen2.5-Math-7B
during training. For the math-specialized model Qwen2.5-Math-7B, the
maximum and mean trajectory lengths decrease substantially over the
course of training, dropping from approximately 3000/1000 tokens at
initialization to around 1500/700 tokens after 1k training iterations,
where they subsequently stabilize. In contrast, for the general-purpose
model Qwen2.5-7B-Instruct, the corresponding values increase from
roughly 1000/500 tokens at the beginning of training to approximately
1800/1000 tokens by the end of training.

These observations suggest that the math-specialized model is initially
biased toward producing excessively long reasoning trajectories, likely
due to pretraining or post-training procedures emphasizing extended
chain-of-thought generation. Our method appears to improve reasoning
efficiency by encouraging shorter yet more effective reasoning chains
that achieve higher accuracy. Conversely, the general-purpose model
is not inherently optimized for long-form mathematical reasoning and
initially produces relatively short solutions. In this case, our method
encourages the generation of longer reasoning trajectories, which
correlates with improved problem-solving performance.

\subsection{Analysis of the Importance Weight in Off-Policy Learning\label{subsec:Analysis-of-the-Importance-Weight}}

\begin{figure}
\begin{centering}
\begin{tabular}{c}
\includegraphics[width=0.99\textwidth]{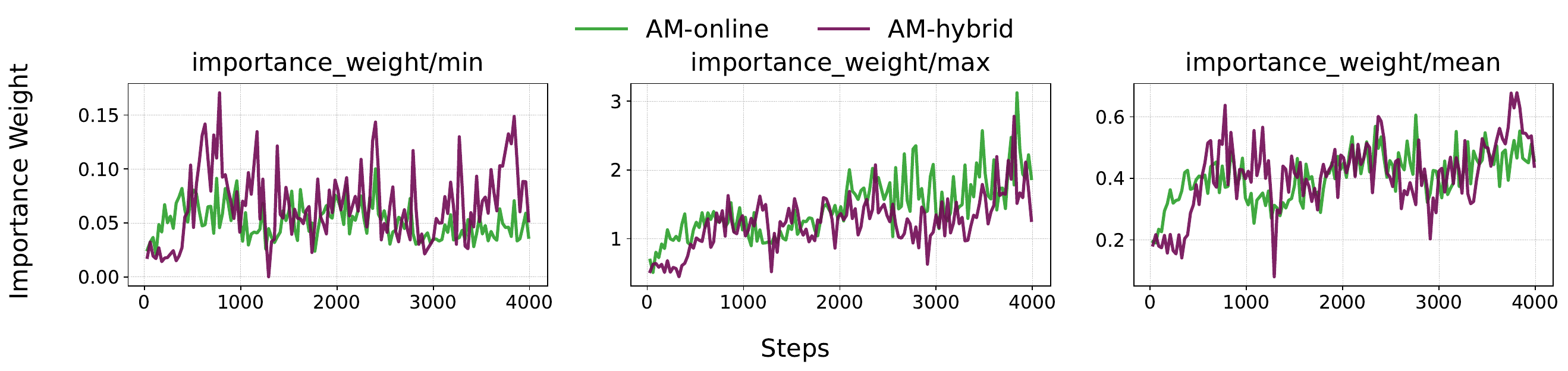}\tabularnewline
(a) Qwen2.5-7B-Instruct\tabularnewline
\tabularnewline
\includegraphics[width=0.99\textwidth]{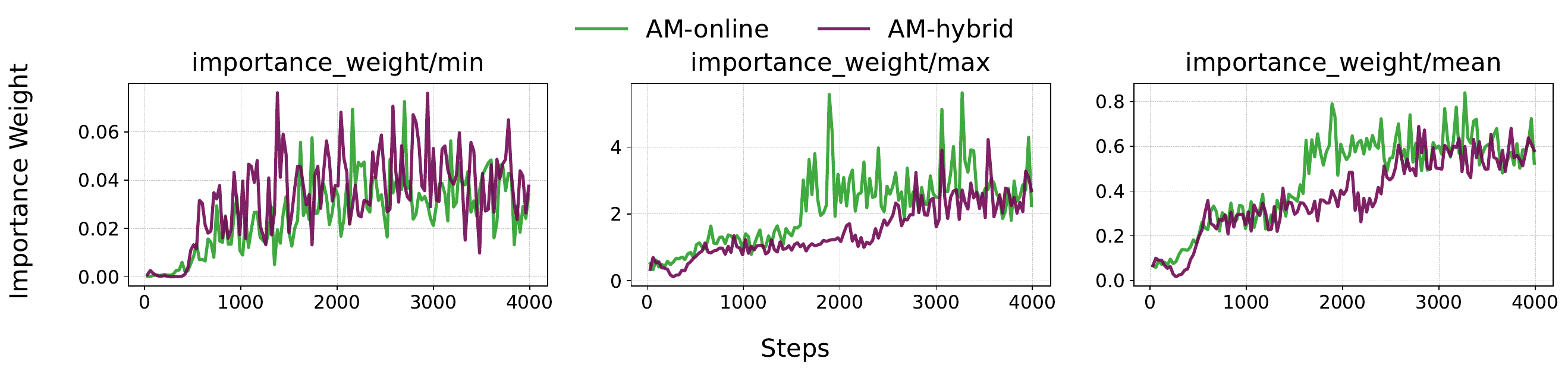}\tabularnewline
(b) Qwen2.5-Math-7B\tabularnewline
\end{tabular}
\par\end{centering}
\caption{Minimum, mean, and maximum trajectories of the importance weights
$w_{\bar{\theta}}\left(s_{0},\tau\right)=\frac{\pi_{\bar{\theta}}\left(\tau|s_{0}\right)}{\pi_{\text{s}}\left(\tau|s_{0}\right)}$
used in $\protect\Loss^{\text{off-policy}}_{\text{AM-Traj}}$(Eq.~\ref{eq:AM_Offpolicy_2})
during training.\label{fig:importance_weight}}
\end{figure}

Figure~\ref{fig:importance_weight} illustrates the minimum, mean,
and maximum values of the trajectory-level importance weight $w_{\bar{\theta}}\left(s_{0},\tau\right)$,
defined as the ratio between the current learned policy and the sampling
policy (served by vLLM) over an entire trajectory. The mean importance
weights deviate from 1, indicating a mismatch between the learned
and sampling policies that introduces sampling bias.

\subsection{Generation Examples}

In this section, we provide representative examples of solutions generated
by our method and baseline approaches for both mathematical reasoning
and code generation. Correct and wrong answers are highlighted in
\textcolor{blue}{blue} and \textcolor{red}{red}, respectively.

\subsubsection{Math Reasoning}

\begin{tabular}{|ll|}
\hline 
\textbf{Dataset:} & Minerva Math\tabularnewline
\textbf{Problem:} & Find a purely sinusoidal solution of $\frac{d^{4}x}{dt^{4}}-x=\cos(2t)$.\tabularnewline
\textbf{GT Answer:} & $\frac{\cos(2t)}{15}$\tabularnewline
\textbf{Ref. Model:} & Qwen2.5-Math-7B\tabularnewline
\hline 
\end{tabular}

\subsubsection*{Base model}

\textbf{\emph{Answer:}} \textcolor{red}{null}

\textbf{\emph{Solution:}}\\
First, we need to find the complementary solution by solving the homogeneous
equation $\frac{d^{4}x}{dt^{4}}-x=0$. The characteristic equation
is $r^{4}-1=0$. Solving this, we get the roots $r=1,-1,i,-i$. Therefore,
the complementary solution is $x_{c}(t)=c_{1}e^{t}+c_{2}e^{-t}+c_{3}\cos(t)+c_{4}\sin(t)$.</think>\\
Next, we need to find the particular solution. Since the right-hand
side is $\cos(2t)$, we assume a particular solution of the form $x_{p}(t)=A\cos(2t)+B\sin(2t)$.
Substituting this into the differential equation and solving for $A$
and $B$, we get the particular solution.\\
Finally, the general solution is the sum of the complementary and
particular solutions.</think>\\
The final answer is <answer> $x(t)=c_{1}e^{t}+c_{2}e^{-t}+c_{3}\cos(t)+c_{4}\sin(t)+\frac{1}{3}\cos(2t)$
</answer>.

\subsubsection*{AM-online}

\textbf{\emph{Answer:}} \textcolor{blue}{$\frac{1}{15}\cos(2t)$}

\textbf{\emph{Solution:}}\\
First, we need to find the general solution to the homogeneous equation
$\frac{d^{4}x}{dt^{4}}-x=0$. The characteristic equation for this
differential equation is $r^{4}-1=0$. The roots of this equation
are $r=1,-1,i,-i$. Therefore, the general solution to the homogeneous
equation is:\\
\[
x_{h}(t)=C_{1}e^{t}+C_{2}e^{-t}+C_{3}\cos(t)+C_{4}\sin(t)
\]

Next, we need to find a particular solution to the non-homogeneous
equation $\frac{d^{4}x}{dt^{4}}-x=\cos(2t)$. Since the right-hand
side is $\cos(2t)$, we assume a particular solution of the form $x_{p}(t)=A\cos(2t)+B\sin(2t)$.
We need to find the values of $A$ and $B$.

Taking the derivatives:\\
\[
x_{p}'(t)=-2A\sin(2t)+2B\cos(2t)
\]
\[
x_{p}''(t)=-4A\cos(2t)-4B\sin(2t)
\]
\[
x_{p}'''(t)=8A\sin(2t)-8B\cos(2t)
\]
\[
x^{(4)}_{p}(t)=16A\cos(2t)+16B\sin(2t)
\]

Substituting $x^{(4)}_{p}(t)$ and $x_{p}(t)$ into the differential
equation:\\
\[
16A\cos(2t)+16B\sin(2t)-(A\cos(2t)+B\sin(2t))=\cos(2t)
\]
\[
(16A-A)\cos(2t)+(16B-B)\sin(2t)=\cos(2t)
\]
\[
15A\cos(2t)+15B\sin(2t)=\cos(2t)
\]

By comparing the coefficients of $\cos(2t)$ and $\sin(2t)$, we get:\\
\[
15A=1\implies A=\frac{1}{15}
\]
\[
15B=0\implies B=0
\]

So the particular solution is:\\
\[
x_{p}(t)=\frac{1}{15}\cos(2t)
\]

The general solution to the differential equation is the sum of the
homogeneous solution and the particular solution:\\
\[
x(t)=C_{1}e^{t}+C_{2}e^{-t}+C_{3}\cos(t)+C_{4}\sin(t)+\frac{1}{15}\cos(2t)
\]

Since we are asked to find a purely sinusoidal solution, we can ignore
the terms $C_{1}e^{t}$ and $C_{2}e^{-t}$, as they are not purely
sinusoidal. Therefore, the purely sinusoidal solution is:\\
\[
x(t)=C_{3}\cos(t)+C_{4}\sin(t)+\frac{1}{15}\cos(2t)
\]

However, if we are looking for a solution that is purely a single
sinusoidal term, we can see that $\frac{1}{15}\cos(2t)$ is the only
purely sinusoidal term in the solution. So the purely sinusoidal solution
is\\
\[
x(t)=\frac{1}{15}\cos(2t)
\]
</think>

The final answer is:\\
\[
\boxed{\frac{1}{15}\cos(2t)}
\]

\subsubsection*{AM-hybrid}

\textbf{\emph{Answer:}} \textcolor{blue}{$\frac{1}{15}\cos(2t)$}

\textbf{\emph{Solution:}}\\
To find a purely sinusoidal solution of the differential equation
$\frac{d^{4}x}{dt^{4}}-x=\cos(2t)$, we will follow these steps:\\
1. {*}{*}Find the complementary solution{*}{*}: First, solve the homogeneous
equation $\frac{d^{4}x}{dt^{4}}-x=0$. The characteristic equation
is $r^{4}-1=0$, which factors as $(r^{2}-1)(r^{2}+1)=0$. The roots
are $r=\pm1,\pm i$. Therefore, the complementary solution is:\\
\[
x_{c}(t)=C_{1}e^{t}+C_{2}e^{-t}+C_{3}\cos(t)+C_{4}\sin(t).
\]
2. {*}{*}Find the particular solution{*}{*}: Since the non-homogeneous
term is $\cos(2t)$, we assume a particular solution of the form $x_{p}(t)=A\cos(2t)+B\sin(2t)$.
We need to find the coefficients $A$ and $B$. First, compute the
derivatives of $x_{p}(t)$:\\
\[
x_{p}'(t)=-2A\sin(2t)+2B\cos(2t),
\]
\[
x_{p}''(t)=-4A\cos(2t)-4B\sin(2t),
\]
\[
x_{p}'''(t)=8A\sin(2t)-8B\cos(2t),
\]
\[
x^{(4)}_{p}(t)=16A\cos(2t)+16B\sin(2t).
\]
Substitute $x_{p}(t)$ and its fourth derivative into the original
differential equation:\\
\[
16A\cos(2t)+16B\sin(2t)-(A\cos(2t)+B\sin(2t))=\cos(2t).
\]
Simplify the left-hand side:\\
\[
(16A-A)\cos(2t)+(16B-B)\sin(2t)=15A\cos(2t)+15B\sin(2t)=\cos(2t).
\]
By comparing coefficients, we get:\\
\[
15A=1\quad\text{and}\quad15B=0.
\]
Solving these equations, we find $A=\frac{1}{15}$ and $B=0$. Therefore,
the particular solution is:\\
\[
x_{p}(t)=\frac{1}{15}\cos(2t).
\]
3. {*}{*}Combine the solutions{*}{*}: The general solution to the
differential equation is the sum of the complementary solution and
the particular solution:\\
\[
x(t)=C_{1}e^{t}+C_{2}e^{-t}+C_{3}\cos(t)+C_{4}\sin(t)+\frac{1}{15}\cos(2t).
\]
Since we are asked to find a purely sinusoidal solution, we discard
the terms involving $e^{t}$ and $e^{-t}$ and the term involving
$\cos(t)$ and $\sin(t)$ (as they are not purely sinusoidal with
frequency 2). The purely sinusoidal solution is:\\
\[
x(t)=\frac{1}{15}\cos(2t).
\]
The final answer is:\\
\[
\boxed{\frac{1}{15}\cos(2t)}
\]
</think>\\
The final answer is $\boxed{\frac{1}{15}\cos(2t)}$.

\subsubsection*{GRPO}

\textbf{\emph{Answer:}} \textcolor{red}{$C_{1}e^{t}+C_{2}e^{-t}+C_{3}\cos(t)+C_{4}\sin(t)-\frac{1}{16}t\sin(2t)$}

\textbf{\emph{Solution:}}\\
First, we need to find the general solution to the homogeneous equation
$\frac{d^{4}x}{dt^{4}}-x=0$. The characteristic equation is $r^{4}-1=0$,
which has roots $r=1,-1,i,-i$. Therefore, the general solution to
the homogeneous equation is:\\
\[
x_{h}(t)=C_{1}e^{t}+C_{2}e^{-t}+C_{3}\cos(t)+C_{4}\sin(t)
\]
Next, we need to find a particular solution to the non-homogeneous
equation $\frac{d^{4}x}{dt^{4}}-x=\cos(2t)$. Since $\cos(2t)$ is
a solution to the homogeneous equation, we assume a particular solution
of the form $x_{p}(t)=At\cos(2t)+Bt\sin(2t)$. We then find the fourth
derivative of $x_{p}(t)$:\\
\[
x_{p}(t)=At\cos(2t)+Bt\sin(2t)
\]
\[
x_{p}'(t)=A\cos(2t)-2At\sin(2t)+B\sin(2t)+2Bt\cos(2t)
\]
\[
x_{p}''(t)=-4At\cos(2t)-4Bt\sin(2t)-4A\sin(2t)+4B\cos(2t)
\]
\[
x_{p}'''(t)=-8At\sin(2t)+8Bt\cos(2t)-8A\cos(2t)-8B\sin(2t)
\]
\[
x^{(4)}_{p}(t)=16At\cos(2t)+16Bt\sin(2t)+16A\sin(2t)-16B\cos(2t)
\]
Substituting $x^{(4)}_{p}(t)$ and $x_{p}(t)$ into the original equation,
we get:\\
\[
16At\cos(2t)+16Bt\sin(2t)+16A\sin(2t)-16B\cos(2t)-At\cos(2t)-Bt\sin(2t)=\cos(2t)
\]
Simplifying, we get:\\
\[
15At\cos(2t)+15Bt\sin(2t)+16A\sin(2t)-16B\cos(2t)=\cos(2t)
\]
By comparing coefficients, we get:\\
\[
15A=0,\quad15B=0,\quad16A=0,\quad-16B=1
\]
Solving these equations, we get $A=0$ and $B=-\frac{1}{16}$. Therefore,
the particular solution is:\\
\[
x_{p}(t)=-\frac{1}{16}t\sin(2t)
\]
The general solution to the non-homogeneous equation is:\\
\[
x(t)=C_{1}e^{t}+C_{2}e^{-t}+C_{3}\cos(t)+C_{4}\sin(t)-\frac{1}{16}t\sin(2t)
\]
</think>\\
The final answer is:\\
\[
\boxed{C_{1}e^{t}+C_{2}e^{-t}+C_{3}\cos(t)+C_{4}\sin(t)-\frac{1}{16}t\sin(2t)}
\]

\subsubsection*{GSPO}

\textbf{\emph{Answer:}} \textcolor{red}{$C_{1}e^{t}+C_{2}e^{-t}+C_{3}\cos(t)+C_{4}\sin(t)+\frac{1}{15}\cos(2t)$}

\textbf{\emph{Solution:}}\\
First, we need to find the complementary solution to the homogeneous
equation $\frac{d^{4}x}{dt^{4}}-x=0$. The characteristic equation
is $r^{4}-1=0$, which gives us the roots $r=1,-1,i,-i$. So the complementary
solution is $x_{c}(t)=C_{1}e^{t}+C_{2}e^{-t}+C_{3}\cos(t)+C_{4}\sin(t)$.\\
Next, we need to find a particular solution to the non-homogeneous
equation $\frac{d^{4}x}{dt^{4}}-x=\cos(2t)$. Since $\cos(2t)$ is
a solution to the homogeneous equation, we assume a particular solution
of the form $x_{p}(t)=A\cos(2t)+B\sin(2t)$. Substituting this into
the differential equation, we get:\\
\[
\frac{d^{4}}{dt^{4}}(A\cos(2t)+B\sin(2t))-(A\cos(2t)+B\sin(2t))=\cos(2t)
\]
\[
16A\cos(2t)+16B\sin(2t)-A\cos(2t)-B\sin(2t)=\cos(2t)
\]
\[
(16A-A)\cos(2t)+(16B-B)\sin(2t)=\cos(2t)
\]
\[
15A\cos(2t)+15B\sin(2t)=\cos(2t)
\]
This gives us the system of equations:\\
\[
15A=1\quad\text{and}\quad15B=0
\]
So, $A=\frac{1}{15}$ and $B=0$. Therefore, the particular solution
is $x_{p}(t)=\frac{1}{15}\cos(2t)$.\\
The general solution is the sum of the complementary solution and
the particular solution:\\
\[
x(t)=C_{1}e^{t}+C_{2}e^{-t}+C_{3}\cos(t)+C_{4}\sin(t)+\frac{1}{15}\cos(2t)
\]
</think>\\
The final answer is $\boxed{C_{1}e^{t}+C_{2}e^{-t}+C_{3}\cos(t)+C_{4}\sin(t)+\frac{1}{15}\cos(2t)}$.

\subsubsection*{Dr.GRPO}

\textbf{\emph{Answer:}}\textbf{ }\textcolor{red}{$C_{3}\cos(t)+C_{4}\sin(t)$}

\textbf{\emph{Solution:}}\\
First, we need to find the complementary solution to the homogeneous
equation $\frac{d^{4}x}{dt^{4}}-x=0$. The characteristic equation
is $r^{4}-1=0$, which has roots $r=1,-1,i,-i$. Therefore, the complementary
solution is:\\
\[
x_{c}(t)=C_{1}e^{t}+C_{2}e^{-t}+C_{3}\cos(t)+C_{4}\sin(t)
\]
Next, we need to find a particular solution to the non-homogeneous
equation $\frac{d^{4}x}{dt^{4}}-x=\cos(2t)$. Since the right-hand
side is $\cos(2t)$, we assume a particular solution of the form $x_{p}(t)=A\cos(2t)+B\sin(2t)$.
Substituting this into the differential equation, we get:\\
\[
\frac{d^{4}}{dt^{4}}(A\cos(2t)+B\sin(2t))-(A\cos(2t)+B\sin(2t))=\cos(2t)
\]
\[
16A\cos(2t)+16B\sin(2t)-A\cos(2t)-B\sin(2t)=\cos(2t)
\]
\[
(16A-A)\cos(2t)+(16B-B)\sin(2t)=\cos(2t)
\]
\[
15A\cos(2t)+15B\sin(2t)=\cos(2t)
\]
By comparing coefficients, we get:\\
\[
15A=1\implies A=\frac{1}{15}
\]
\[
15B=0\implies B=0
\]
So the particular solution is:\\
\[
x_{p}(t)=\frac{1}{15}\cos(2t)
\]
The general solution to the differential equation is the sum of the
complementary solution and the particular solution:\\
\[
x(t)=C_{1}e^{t}+C_{2}e^{-t}+C_{3}\cos(t)+C_{4}\sin(t)+\frac{1}{15}\cos(2t)
\]
Since we are asked to find a purely sinusoidal solution, we can ignore
the terms $C_{1}e^{t}$ and $C_{2}e^{-t}$, as they are not purely
sinusoidal. Therefore, the purely sinusoidal solution is:\\
\[
x(t)=C_{3}\cos(t)+C_{4}\sin(t)+\frac{1}{15}\cos(2t)
\]
However, if we are only interested in the purely sinusoidal part,
we can write:\\
\[
x(t)=C_{3}\cos(t)+C_{4}\sin(t)
\]
</think>\\
The purely sinusoidal solution of the differential equation $\frac{d^{4}x}{dt^{4}}-x=\cos(2t)$
is:\\
\[
\boxed{C_{3}\cos(t)+C_{4}\sin(t)}
\]

\end{document}